\documentclass[12pt]{article}%
\usepackage{amsmath}
\usepackage{amsfonts}
\usepackage{amssymb}
\usepackage{graphicx}%
\providecommand{\U}[1]{\protect\rule{.1in}{.1in}}
\begin{document}

\title{A robust algorithm for explaining unreliable machine learning survival models
using the Kolmogorov-Smirnov bounds}
\author{Maxim S. Kovalev and Lev V. Utkin\\Peter the Great St.Petersburg Polytechnic University (SPbPU)\\St.Petersburg, Russia\\e-mail: lev.utkin@gmail.com, maxkovalev03@gmail.com}
\date{}
\maketitle

\begin{abstract}
A new robust algorithm based of the explanation method SurvLIME called
SurvLIME-KS is proposed for explaining machine learning survival models. The
algorithm is developed to ensure robustness to cases of a small amount of
training data or outliers of survival data. The first idea behind SurvLIME-KS
is to apply the Cox proportional hazards model to approximate the black-box
survival model at the local area around a test example due to the linear
relationship of covariates in the model. The second idea is to incorporate the
well-known Kolmogorov-Smirnov bounds for constructing sets of predicted
cumulative hazard functions. As a result, the robust maximin strategy is used,
which aims to minimize the average distance between cumulative hazard
functions of the explained black-box model and of the approximating Cox model,
and to maximize the distance over all CHFs in the interval produced by the
Kolmogorov-Smirnov bounds. The maximin optimization problem is reduced to the
quadratic program. Various numerical experiments with synthetic and real
datasets demonstrate the SurvLIME-KS efficiency.


\end{abstract}

\section{Introduction}

A rapid and significant success in applying machine learning models to a wide
range of applications, especially to medicine \cite{Holzinger-etal-2019},
meets a problem of understanding results provided by the models, for example,
a doctor has to have an explanation of a stated diagnosis in order to choose a
corresponding treatment. However, many models are black-box, i.e., their
inputs and outcomes may be known for users, but it is not clear what
information in the input data makes them actually arrive at their decisions.
Therefore, explanations of the machine learning model prediction can help
machine users or machine learning experts to understand the obtained results,
and the model explainability becomes to be an important component of machine
learning and even a key factor of decision making
\cite{Arya-etal-2019,Guidotti-2019,Molnar-2019,Murdoch-etal-2019}.

One of the groups of explanation methods consists of the so-called local
methods which derive explanation locally around a test example. They aim to
assign to every feature of the test example some number quantified its impact
on the prediction in order to explain the contribution of individual input
features. One of the first local explanation methods is the Local
Interpretable Model-agnostic Explanations (LIME) \cite{Ribeiro-etal-2016},
which uses simple and easily understandable linear models to locally
approximate the predictions of black-box models. According to LIME, the
explanation may be derived locally from a set of synthetic examples generated
randomly in the neighborhood of the example. A thorough theoretical analysis
of LIME is given by Garreau and Luxburg \cite{Garreau-Luxburg-2020} where the
authors note that LIME is flexible to provide explanations for different data
types, including text and image data.

Among a large variety of machine learning models, we select the models which
try to solve a fundamental problem of survival analysis to understand the
relationship between the covariates and the distribution of survival times to
events of interest. Survival analysis is an important field of statistics
which aims at predicting the time to events of interest, and it can
simultaneously model event data and censored data
\cite{Hosmer-Lemeshow-May-2008}. By taking into account the importance of the
survival analysis tasks, a lot of machine learning survival models have been
developed \cite{Lee-Zame-etal-2018,Wang-Li-Reddy-2017,Zhao-Feng-2019}. One of
the main peculiarities of the survival models is that their outcomes are
functions (the survival function, the hazard function, the cumulative hazard
function, etc.) as predictions instead of points.

A popular model, which establishes the relationship between the covariates and
the distribution of survival times, is the semi-parametric Cox proportional
hazards model \cite{Cox-1972}. The model assumes that a patient's log-risk of
failure is a linear combination of the example covariates. This is a very
important property of the Cox model, which can be used below in explanation models.

Taking into account the need to explain the machine learning black-box
survival models, Kovalev et al. \cite{Kovalev-Utkin-Kasimov-20} proposed an
explanation method called SurvLIME, which deals with censored data and can be
regarded as an extension of LIME on the case of survival data. The basic idea
behind SurvLIME is to apply the Cox model to approximate the black-box
survival model at a local area around a test example. The Cox model is chosen
due to its assumption of the linear combination of covariates. This assumption
implies that coefficients of the covariates can be viewed as quantitative
impacts on the prediction.

The main problem of using SurvLIME is a lack of robustness to cases of a small
amount of training data or outliers of survival data. There are many machine
learning methods which try to cope with this problem. One of the ways to
implement the robust models is to incorporate the imprecise probability or
statistical inference models \cite{Walley91}, which use sets of probability
distributions on data instead of a single distribution. The robust models
based on imprecise probabilities use the maximin strategy, which can be
interpreted as an insurance against the worst case because it aims at
minimizing the expected loss in the least favorable case \cite{Robert94}.
Robust models have been widely exploited in regression and classification
problems due to the opportunity to avoid some strong assumptions underlying
the standard classification models \cite{Xu:2009}. The imprecise probability
models have been applied also to survival models (see, for example,
\cite{Mangili-etal-2015}). However, we have to point out that the above robust
models have been incorporated into the machine learning models themselves, but
not to their explanations. Therefore, our aim is to develop an algorithm for
explaining the machine learning survival models, which could be robust to the
small amount of training data and to outliers in the training or testing sets.

Pursuing this goal, we propose a robust explanation algorithm which is called
as SurvLIME-KS and can be regarded as a modification of SurvLIME. It uses the
well-known Kolmogorov-Smirnov (KS) bounds \cite{Johnson-Leone64} for the
predicted cumulative hazard function (CHF). KS bounds make no distributional
assumptions and can be considered in the framework of the imprecise
statistical models. KS bounds have been used to realize robust machine
learning models (see, for example, \cite{Utkin-Coolen-2014}). The following
ideas are the basis for SurvLIME-KS:

\begin{enumerate}
\item SurvLIME is used as a basis for the proposed robust algorithm. It
generates many points at a local area around a test example, and the CHF is
predicted for every generated example by the black-box machine learning model
which has to be explained.

\item Every CHF predicted by the black-box machine learning model is
transformed to a cumulative distribution function for which KS bounds are
determined in accordance with the predefined confidence probability. In order
to construct bounds for the CHF, the inverse transformation is carried out.

\item A maximin optimization problem is stated and solved for getting optimal
coefficients of the approximating Cox model, where the minimum is defined for
the average distance between logarithms of CHFs produced by the explained
black-box model and by the approximating Cox model, it is taken over
coefficients of the approximating Cox model, the maximum is taken over
logarithms of all CHFs restricted by the obtained bounds. The distance between
two functions is based on $L_{\infty}$-norm.\textbf{ }Chebyshev distance
metrics is used because it has an inferior computational cost due to its
simple formulation.

\item The problem is reduced to a standard quadratic optimization problem with
linear constraints.
\end{enumerate}

Many numerical experiments illustrate SurvLIME-KS under different conditions
of training data. It is shown on synthetic and real data that the algorithm
provides outperforming results for small data and under outliers.

The paper is organized as follows. Related work is in Section 2. Basic
concepts of survival analysis and the Cox model are given in Section 3. A
brief description of the method LIME is provided in Section 4. An introduction
to KS bounds is given in Section 5. Section 6 provides a description of the
proposed algorithm SurvLIME-KS and its basic ideas. A derivation of the
optimization problem for determining important features under condition of the
lack of KS bounds, i.e., when the predicted CHFs are precise, can be found in
Section 7. Questions of incorporating KS bounds into the explanation method
and reducing the maximin optimization problem to the quadratic programming
problem are considered in Section 8. Numerical experiments with synthetic data
and real data are provided in Section 9. Concluding remarks can be found in
Section 10.

\section{Related work}

\textbf{Local explanation methods.} Due to importance of the machine learning
model explanation in many applications, a lot of methods have been proposed to
locally explain black-box models. Following the original LIME
\cite{Ribeiro-etal-2016}, a lot of its modifications have been developed due
to a simple nice idea underlying the method to construct a linear
approximating model in a local area around a test example. These modifications
are ALIME \cite{Shankaranarayana-Runje-2019}, NormLIME \cite{Ahern-etal-2019},
DLIME \cite{Zafar-Khan-2019}, Anchor LIME \cite{Ribeiro-etal-2018}, LIME-SUP
\cite{Hu-Chen-Nair-Sudjianto-2018}, LIME-Aleph \cite{Rabold-etal-2019},
GraphLIME \cite{Huang-Yamada-etal-2020}, SurvLIME
\cite{Kovalev-Utkin-Kasimov-20}. Another explanation method, which is based on
linear approximation, is the SHAP
\cite{Lundberg-Lee-2017,Strumbel-Kononenko-2010}, which takes a game-theoretic
approach for optimizing a regression loss function based on Shapley values.

In order to get intuitive and human-friendly explanations, another explanation
technique called as counterfactual explanations \cite{Wachter-etal-2017} was
developed by several authors
\cite{Goyal-etal-2018,Hendricks-etal-2018,Looveren-Klaise-2019,Waa-etal-2018}.
The corresponding methods tell us what to do in order to achieve a desired
outcome. Counterfactual modifications of the LIME were proposed by Ramon et
al. \cite{Ramon-etal-2020} and White and Garcez \cite{White-Garcez-2020}.

Another important group of explanation methods is based on perturbation
techniques
\cite{Fong-Vedaldi-2019,Fong-Vedaldi-2017,Petsiuk-etal-2018,Vu-etal-2019},
which are also used in LIME. The basic idea behind the perturbation techniques
is that contribution of a feature can be determined by measuring how a
prediction score changes when the feature is altered \cite{Du-Liu-Hu-2019}.
Perturbation techniques can be applied to a black-box model without any need
to access the internal structure of the model. However, the corresponding
methods are computationally complex when samples are of the high dimensionality.

A lot of explanation methods, their analysis and critical review can be found
in survey papers
\cite{Adadi-Berrada-2018,Arrieta-etal-2019,Carvalho-etal-2019,Guidotti-2019,Rudin-2019,Xie-Ras-etal-2020}%
.

Most explanation methods deal with the point-valued results produced by
explainable black-box models, for example, with classes of examples. In
contrast to these models, outcomes of survival models are function, for
example, SFs or CHFs. It follows that LIME should be extended on the case of
models with functional outcomes, in particular, with survival models. An
example of such the extended model is SurvLIME \cite{Kovalev-Utkin-Kasimov-20}.

\textbf{Machine learning models in survival analysis}. Wang et al.
\cite{Wang-Li-Reddy-2017} provided a comprehensive review of the machine
learning models dealing with survival analysis problems. The most powerful and
popular method for dealing with survival data is the Cox model \cite{Cox-1972}%
. Following this model, a lot of its modifications have been developed in
order to relax some strong assumption underlying the Cox model. In order to
take into account the high dimensionality of survival data and to solve the
feature selection problem with these data, Tibshirani \cite{Tibshirani-1997}
presented a modification based on the Lasso method. Similar Lasso
modifications, for example, the adaptive Lasso, were also proposed by several
authors \cite{Kim-etal-2012,Witten-Tibshirani-2010,Zhang-Lu-2007}. The next
extension of the Cox model is a set of SVM modifications
\cite{Khan-Zubek-2008,Widodo-Yang-2011}. Various architectures of neural
networks, starting from a simple network \cite{Faraggi-Simon-1995} proposed to
relax the linear relationship assumption in the Cox model, have been developed
\cite{Haarburger-etal-2018,Katzman-etal-2018,Ranganath-etal-2016,Zhu-Yao-Huang-2016}
to solve prediction problems in the framework of survival analysis. In spite
of many powerful machine learning approaches for solving the survival
problems, the most efficient and popular tool for survival analysis under
condition of small survival data is the extension of the standard random
forest \cite{Breiman-2001} called the random survival forest (RSF)
\cite{Ibrahim-etal-2008,Mogensen-etal-2012,Wang-Zhou-2017,Wright-etal-2017}.

Most of the above models dealing with survival data can be regarded as
black-box models and should be explained. However, only the Cox model has a
simple explanation due to its linear relationship between covariates.
Therefore, it can be used to approximate more powerful models, including
survival deep neural networks and random survival forests, in order to explain
predictions of these models.

\textbf{Imprecise probabilities in classification and regression}. There are a
lot of results devoted to application of the imprecise probability models
\cite{Walley96a} to classification and regression problems. Several authors
\cite{Corani-Zaffalon08,Zaffalon-2005} proposed \textquotedblleft
imprecise\textquotedblright\ classifiers that are reliable even in the
presence of small sample sizes and missing values due to use of imprecise
statistical models. Modifications of SVM and random forests on the basis of
incorporating the imprecise models have been presented in papers
\cite{Utkin-14a,Utkin-2019,Utkin-2020}.

Robust classification and regression models using the Kolmogorov-Smirnov
bounds have been applied to classification \cite{Utkin-Coolen-2014} and
regression problems \cite{Utkin-Coolen-2011,Utkin-Wiencierz-2015} to take into
account the lack of sufficient training data and for constructing accurate
classification or regression models. One of the first ideas of applying
imprecise probability models to classification decision trees was presented in
\cite{Abellan-Moral-2003}, where probabilities of classes at decision tree
leaves are estimated by using an imprecise model. Following this work, several
papers devoted to applications of imprecise probabilities to decision trees
and random forests were proposed
\cite{Abellan-etal-2014,Abellan-etal-2017,Abellan-etal-2018,Mantas-Abellan-2014,Moral-Garcia-etal-2020}%
, where the authors developed new splitting criteria taking into account
imprecision of training data and noisy data. Imprecise probabilities have also
been used in classification problems in
\cite{Destercke-Antoine-2013,Matt-2017,Moral-2019}.

\textbf{Imprecise probabilities in survival analysis.} One of the ways for
getting robust survival models is to incorporate imprecise probability models
into survival analysis.\textbf{ }Mangili et al. \cite{Mangili-etal-2015}
proposed a robust estimation of survival functions from right censored data by
introducing a robust Dirichlet process. In this paper, special bounds for SFs
and the corresponding survival curve estimator are presented.

Another approach illustrating how the imprecise Dirichlet model can be used to
determine upper and lower values on the survival function was proposed in
\cite{Bickis-2009,Bickis-Bickis-2007}. Application of the imprecise Dirichlet
model to survival data was also studied by Coolen \cite{Coolen-97}. Coolen and
Yan \cite{Coolen-Yan-2004} considered a generalization of Hill's assumption,
which is used for prediction in case of extremely vague prior knowledge about
the underlying distribution, for dealing with right-censored observations.
Modifications of random survival forests by assigning weights to decision
trees in a way that allows us to control the imprecision and to implement the
robust strategy of decision making are proposed in
\cite{Utkin-Kovalev-Coolen-2019,Utkin-Kovalev-Meldo-Coolen-2019}.

In spite of many papers devoted to the robust machine learning models
mentioned above, there are no methods of the robust local explanation of
survival models. Therefore, the proposed incorporating KS bounds into SurvLIME
can be regarded as the first robust explanation method of the survival model predictions.

\section{Some elements of survival analysis}

\subsection{Basic concepts}

In survival analysis, an example (patient) $i$ is represented by a triplet
$(\mathbf{x}_{i},\delta_{i},T_{i})$, where $\mathbf{x}_{i}=(x_{i1}%
,...,x_{id})$ is the vector of the patient parameters (characteristics) or the
vector of the example features; $T_{i}$ is time to event of the example. If
the event of interest is observed, $T_{i}$ corresponds to the time between
baseline time and the time of event happening, in this case $\delta_{i}=1$,
and we have an uncensored observation. If the example event is not observed
and its time to event is greater than the observation time, $T_{i}$
corresponds to the time between baseline time and end of the observation, and
the event indicator is $\delta_{i}=0$, and we have a censored observation.
Suppose a training set $D$ consists of $n$ triplets $(\mathbf{x}_{i}%
,\delta_{i},T_{i})$, $i=1,...,n$. The goal of survival analysis is to estimate
the time to the event of interest $T$ for a new example (patient) with feature
vector denoted by $\mathbf{x}$ by using the training set $D$.

The survival and hazard functions are key concepts in survival analysis for
describing the distribution of event times. The survival function (SF) denoted
by $S(t|\mathbf{x})$ as a function of time $t$ is the probability of surviving
up to that time, i.e., $S(t|\mathbf{x})=\Pr\{T>t|\mathbf{x}\}$. The hazard
function $h(t|\mathbf{x})$ is the rate of event at time $t$ given that no
event occurred before time $t$, i.e., $h(t|\mathbf{x})=f(t|\mathbf{x}%
)/S(t|\mathbf{x})$, where $f(t|\mathbf{x})$ is the density function of the
event of interest. The hazard rate is defined as
\begin{equation}
h(t|\mathbf{x})=-\frac{\mathrm{d}}{\mathrm{d}t}\ln S(t|\mathbf{x}).
\end{equation}

Another important concept is the CHF $H(t|\mathbf{x})$, which is defined as
the integral of the hazard function $h(t|\mathbf{x})$ and can be interpreted
as the probability of an event at time $t$ given survival until time $t$,
i.e.,
\begin{equation}
H(t|\mathbf{x})=\int_{-\infty}^{t}h(x|\mathbf{x})dx.
\end{equation}

The SF can be expressed through the CHF as $S(t|\mathbf{x})=\exp\left(
-H(t|\mathbf{x})\right)  $.

To compare the survival models, the C-index proposed by Harrell et al.
\cite{Harrell-etal-1982} is used. It estimates how good the model is at
ranking survival times. It estimates the probability that, in a randomly
selected pair of examples, the example that fails first had a worst predicted
outcome. In fact, this is the probability that the event times of a pair of
examples are correctly ranking.

\subsection{The Cox model}

Let us consider main concepts of the Cox proportional hazards model,
\cite{Hosmer-Lemeshow-May-2008}. According to the model, the hazard function
at time $t$ given predictor values $\mathbf{x}$ is defined as
\begin{equation}
h(t|\mathbf{x},\mathbf{b})=h_{0}(t)\Psi(\mathbf{x},\mathbf{b})=h_{0}%
(t)\exp\left(  \psi(\mathbf{x},\mathbf{b})\right)  .
\end{equation}

Here $h_{0}(t)$ is a baseline hazard function which does not depend on the
vector $\mathbf{x}$ and the vector $\mathbf{b}$; $\Psi(\mathbf{x})$ is the
covariate effect or the risk function; $\mathbf{b}=(b_{1},...,b_{d})$ is an
unknown vector of regression coefficients or parameters. It can be seen from
the above expression for the hazard function that the reparametrization
$\Psi(\mathbf{x},\mathbf{b})=\exp\left(  \psi(\mathbf{x},\mathbf{b})\right)  $
is used in the Cox model. The function $\psi(\mathbf{x},\mathbf{b})$ in the
model is linear, i.e.,
\begin{equation}
\psi(\mathbf{x},\mathbf{b})=\mathbf{xb}^{\mathrm{T}}=\sum\nolimits_{k=1}%
^{d}b_{k}x_{k}.
\end{equation}

In the framework of the Cox model, the survival function $S(t|\mathbf{x}%
,\mathbf{b})$ is computed as
\begin{equation}
S(t|\mathbf{x},\mathbf{b})=\exp(-H_{0}(t)\exp\left(  \psi(\mathbf{x}%
,\mathbf{b})\right)  )=\left(  S_{0}(t)\right)  ^{\exp\left(  \psi
(\mathbf{x},\mathbf{b})\right)  }.
\end{equation}

Here $H_{0}(t)$ is the cumulative baseline hazard function; $S_{0}(t)$ is the
baseline survival function. It is important to note that functions $H_{0}(t)$
and $S_{0}(t)$ do not depend on $\mathbf{x}$ and $\mathbf{b}$.

The partial likelihood in this case is defined as follows:
\begin{equation}
L(\mathbf{b})=\prod_{j=1}^{n}\left[  \frac{\exp(\psi(\mathbf{x}_{j}%
,\mathbf{b}))}{\sum_{i\in R_{j}}\exp(\psi(\mathbf{x}_{i},\mathbf{b}))}\right]
^{\delta_{j}}.
\end{equation}

Here $R_{j}$ is the set of patients who are at risk at time $t_{j}$. The term
\textquotedblleft at risk at time $t$\textquotedblright\ means patients who
die at time $t$ or later.

\section{LIME}

Let us briefly consider the original LIME \cite{Ribeiro-etal-2016}. It is an
explanation framework for the decision of many machine learning classifiers.
The method proposes to approximate a black-box explainable model denoted as
$f$ with a simple function $g$ in the vicinity of the point of interest
$\mathbf{x}$, whose prediction by means of $f$ has to be explained, under
condition that the approximation function $g$ belongs to a set of explanation
models $G$, for example, linear models. In order to construct the function $g$
in accordance with LIME, a new dataset consisting of perturbed samples is
generated, and predictions corresponding to the perturbed samples are obtained
by means of the explained model. New samples are assigned by weights
$w_{\mathbf{x}}$ in accordance with their proximity to the point of interest
$\mathbf{x}$ by using a distance metric, for example, the Euclidean distance.

An explanation (local surrogate) model is trained on new generated samples by
solving the following optimization problem:
\begin{equation}
\arg\min_{g\in G}L(f,g,w_{\mathbf{x}})+\Phi(g).
\end{equation}

Here $L$ is a loss function, for example, mean squared error, which measures
how the explanation is close to the prediction of the explainable model;
$\Phi(g)$ is the model complexity.

Finally, LIME provides a local linear model which explains the prediction by
analyzing its coefficients.

\section{Kolmogorov-Smirnov bounds}

One of the ways for taking into account the amount of statistical data and for
constructing bounds for the set of probability distributions is using the
Kolmogorov-Smirnov confidence limits for the empirical cumulative distribution
function $F_{n}(\mathbf{x})$ constructed on the basis of $n$ observations.

Suppose that function $F(\mathbf{x})$ is a true probability distribution
function of points from the training set. It is assumed that $F(\mathbf{x})$
is unknown. If the training set consists of $n$ examples, then a critical
value of the test statistic $d_{n,1-\gamma}$ can be chosen such that a band of
width $\pm d_{n,1-\gamma}$ around $F_{n}(\mathbf{x})$ will entirely contain
$F(\mathbf{x})$ with probability $1-\gamma$, which is to be interpreted as a
confidence statement in the frequentist statistical framework. In other words,
we can write $\Pr\{D_{n}\geq d_{n,1-\gamma}\}=\gamma$, where the quantity
$D_{n}=\max_{\mathbf{x}}\left\vert F_{n}(\mathbf{x})-F(\mathbf{x})\right\vert
$ is called the Kolmogorov-Smirnov statistic. Denote the $(1-\gamma)$-quantile
of the Kolmogorov distribution by $k_{1-\gamma}$. The ways for computing
$d_{n,1-\gamma}$ for given $n$ and $\gamma$ as well as the values of
$k_{1-\gamma}$ can be found in the book \cite{Johnson-Leone64}. In particular,
according to \cite{Johnson-Leone64}, a good approximation for the test
statistic for $n>10$ is given by
\begin{equation}
d_{n,1-\gamma}\approx k_{1-\gamma}/\sqrt{n}.
\end{equation}
For $n\leq10$ another approximation can be used \cite{Johnson-Leone64}:%
\begin{equation}
d_{n,1-\gamma}\approx k_{1-\gamma}\left(  \sqrt{n}+0.12+0.11/\sqrt{n}\right)
^{-1}.
\end{equation}

Taking into account that the bounds are cumulative distribution functions, we
write the following bounds $F_{n}^{L}(\mathbf{x})$ and $F_{n}^{U}(\mathbf{x})$
for some unknown distribution function $F(\mathbf{x})$:
\begin{equation}
F_{n}^{L}(\mathbf{x})\leq F(\mathbf{x})\leq F_{n}^{U}(\mathbf{x}),
\label{NonPar_KS_2}%
\end{equation}
where
\begin{equation}
F_{n}^{L}(\mathbf{x})=\max(F_{n}(\mathbf{x})-d_{n,1-\gamma},0),
\end{equation}%
\begin{equation}
F_{n}^{U}(\mathbf{x})=\min(F_{n}(\mathbf{x})+d_{n,1-\gamma},1).
\end{equation}

It can be seen from the above inequality that the left tail of the upper
probability distribution is $d_{n,1-\gamma}$. The right tail of the lower
probability distribution is $1-d_{n,1-\gamma}$. The corresponding jump is
located at boundary points of the sample space far from all data points (see
\cite{Utkin-Coolen-2011} for details).

It is also important to note that KS bounds depend on the number of training
examples $n$. This implies that these bounds can be applied to the robust
explanation of the survival machine learning model predictions. The bounds
allow us to develop new explanation models which take into account the lack of
sufficient training data.

\section{A sketch of SurvLIME-KS}

SurvLIME-KS can be regarded as an extension of SurvLIME. Therefore, the first
part of the sketch is devoted to SurvLIME. The second part is its extension.

Given a training set $D$ and a black-box model which produces an output in the
form of the CHF $H(t|\mathbf{x})$ for every new example $\mathbf{x}$. An idea
behind SurvLIME is to approximate the output of the black-box model by means
of the Cox model with the CHF denoted as $H_{\text{Cox}}(t|\mathbf{x}%
,\mathbf{b})$ for the same input example $\mathbf{x}$, where parameters
$\mathbf{b}$ are unknown a priori. This approximation allows us to get the
coefficients $\mathbf{b}$ of the approximating Cox model, whose values can be
regarded as quantitative impacts on the prediction $H(t|\mathbf{x})$. The
largest coefficients indicate the corresponding important features.

Optimal coefficients $\mathbf{b}$ make the distance between CHFs
$H(t|\mathbf{x})$ and $H_{\text{Cox}}(t|\mathbf{x},\mathbf{b})$ for the
example $\mathbf{x}$ as small as possible. Similarly to LIME, we consider many
nearest examples $\mathbf{x}_{k}$ generated in a local area around
$\mathbf{x}$. For every generated $\mathbf{x}_{k}$, the CHF $H(t|\mathbf{x}%
_{k})$ of the black-box model is predicted. By taking into account generated
points and the corresponding CHFs $H(t|\mathbf{x}_{k})$, optimal values of
$\mathbf{b}$ minimize the weighted average distance between every pair of CHFs
$H(t|\mathbf{x}_{k})$ and $H_{\text{Cox}}(t|\mathbf{x}_{k},\mathbf{b})$ over
all points $\mathbf{x}_{k}$. Weight $w_{k}$ is assigned to the distance
between points $\mathbf{x}_{k}$ and $\mathbf{x}$ in accordance with its value.
In particular, smaller distances between $\mathbf{x}_{k}$ and $\mathbf{x}$
produce larger weights of distances between CHFs.

The distance metric between CHFs $H(t|\mathbf{x}_{k})$ and $H_{\text{Cox}%
}(t|\mathbf{x}_{k},\mathbf{b})$ defines the corresponding optimization problem
for computing optimal coefficients $\mathbf{b}$. One of the possible ways is
to apply $L_{p}$-norms as a measure of a distance between two function.
However, the direct use of $L_{p}$-norms for defining distances between CHFs
$H(t|\mathbf{x}_{k})$ and $H_{\text{Cox}}(t|\mathbf{x}_{k},\mathbf{b})$ may
lead to extremely complex optimization problems. Therefore, SurvLIME
\cite{Kovalev-Utkin-Kasimov-20} uses the $L_{2}$-norm applied to logarithms of
CHFs $H(t|\mathbf{x}_{k})$ and $H_{\text{Cox}}(t|\mathbf{x}_{k},\mathbf{b})$,
which leads to a convex optimization problem.

In contrast to SurvLIME, SurvLIME-KS uses the $L_{\infty}$-norm as a distance
between logarithms of CHFs $H(t|\mathbf{x}_{k})$ and $H_{\text{Cox}%
}(t|\mathbf{x}_{k},\mathbf{b})$. It leads to a simpler optimization problem
for computing the vector $\mathbf{b}$.

The first idea behind the SurvLIME-KS implementation is to define bounds for
CHFs $H(t|\mathbf{x}_{k})$ provided by the black-box model and their
logarithms $\phi(t|\mathbf{x}_{k})=\ln H(t|\mathbf{x}_{k})$ for all generated
points $\mathbf{x}_{k}$. The logarithm is taken to simplify the optimization
problem. The second idea is to extend the minimization problem for computing
coefficients $\mathbf{b}$ on a maximin optimization problem, where the maximum
is taken over all logarithms of CHFs $H(t|\mathbf{x}_{k})$ restricted by the
defined bounds. The problem is solved by representing the minimization problem
by the dual one. As a result, we get the maximization problem over logarithms
of CHFs and coefficients $\mathbf{b}$. By adding the regularization term in
the form of the quadratic norm, we get the standard quadratic optimization
problem with linear constraints.

\begin{figure}
[ptb]
\begin{center}
\includegraphics[
%%=19.291500in,
%%=29.145800in,
height=3.3797in,
width=5.0903in
]%
{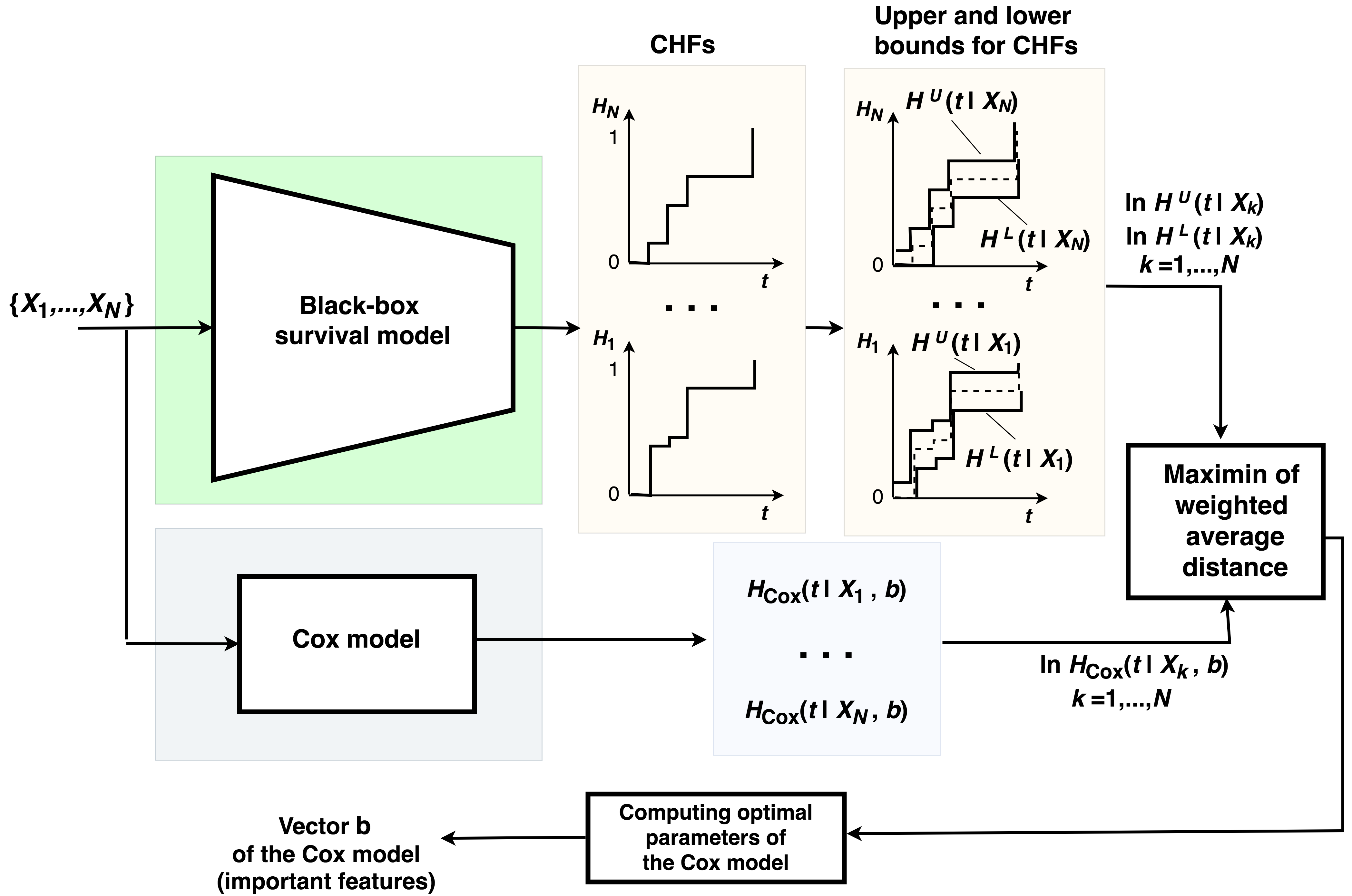}%
\caption{A schematic illustration of the explanation algorithm}%
\label{fig:chf_ks_explain_1}%
\end{center}
\end{figure}

Fig. \ref{fig:chf_ks_explain_1} illustrates the explanation algorithm. It can
be seen from Fig. \ref{fig:chf_ks_explain_1} that a set of examples
$\{\mathbf{x}_{1},...,\mathbf{x}_{N}\}$ are fed to the black-box survival
model, which produces a set of CHFs $\{H(t|\mathbf{x}_{1}),...,H(t|\mathbf{x}%
_{N})\}$. Simultaneously, we write CHFs $H_{\text{Cox}}(t|\mathbf{x}%
_{k},\mathbf{b})$, $k=1,...,N$, as functions of coefficients $\mathbf{b}$ for
all generated examples. Then for every obtained CHF $H(t|\mathbf{x}_{k})$, the
lower $H^{L}(t|\mathbf{x}_{k})$ and upper $H^{U}(t|\mathbf{x}_{k})$ bounds are
introduced by using Kolmogorov-Smirnov bounds. The maximin optimization
problem is written with the objective function which maximizes minimal
weighted average distance between logarithms of CHFs $\ln H_{\text{Cox}%
}(t|\mathbf{x}_{k},\mathbf{b})$ and $\ln H(t|\mathbf{x}_{k})$ from a set of
functions restricted by boundary functions $\ln H^{L}(t|\mathbf{x}_{k})$ and
$\ln H^{U}(t|\mathbf{x}_{k})$. Optimal values of coefficients $\mathbf{b}$ is
the solution to the problem.%

%TCIMACRO{\FRAME{ftbpFU}{5.0903in}{3.3797in}{0pt}{\Qcb{A schematic illustration
%of the explanation algorithm}}{\Qlb{fig:chf_ks_explain_1}}%
%{chf_ks_explain_1.png}{\special{ language "Scientific Word";  type "GRAPHIC";
%width 5.0903in;  height 3.3797in;  depth 0pt;  original-width 29.1458in;
%original-height 19.2915in;  cropleft "0";  croptop "1";  cropright "1";
%cropbottom "0";  filename 'CHF_KS_explain_1.png';file-properties "XNPEU";}} }%
%BeginExpansion
%EndExpansion

\section{$L_{\infty}$-norm and the first part of SurvLIME-KS}

In accordance with the sketch of the algorithm, $N$ points $\mathbf{x}_{k}$,
$k=1,...,N$, are generated in a local area around $\mathbf{x}$. Let us find
the distances between CHFs provided by the black-box model for generated
points $\mathbf{x}_{k}$, $k=1,...,N$, and the CHF provided by the
approximating Cox model for point $\mathbf{x}$. Before deriving the distances,
we introduce some notations and conditions.

Let $t_{0}<t_{1}<...<t_{m}$ be the distinct times to event of interest from
the set $\{T_{1},...,T_{n}\}$, where $t_{0}=\min_{k=1,...,n}T_{k}$ and
$t_{m}=\max_{k=1,...,n}T_{k}$. The black-box model maps the feature vectors
$\mathbf{x}\in\mathbb{R}^{d}$ into piecewise linear CHFs $H(t|\mathbf{x})$
such that $H(t|\mathbf{x})\geq0$ for all $t$ and $\max_{t}H(t|\mathbf{x}%
)<\infty$. Let us introduce the time $T=t_{m}+\gamma$ in order to restrict the
function $H(t|\mathbf{x})$, where $\gamma$ is a very small positive number.
Let $\Omega=[0,T]$ and divide it into $m+1$ subsets $\Omega_{0},...,\Omega
_{m}$ such that $\Omega=\cup_{j=0,...,m}\Omega_{j}$; $\Omega_{m}=[t_{m},T]$,
$\Omega_{j}=[t_{j},t_{j+1})$, $\forall j\in\{0,...,m-1\}$; $\Omega_{j}%
\cap\Omega_{k}=\emptyset$, $\forall j\neq k$.

Since the CHF $H(t|\mathbf{x})$ is piecewise constant, then it can be written
as
\begin{equation}
H(t|\mathbf{x})=\sum_{j=0}^{m}H_{j}(\mathbf{x})\cdot\chi_{j}(t)
\label{KS_Inf_SurvLIME_20}%
\end{equation}
under additional condition $H_{j}(\mathbf{x})\geq\varepsilon>0$, where
$\varepsilon$ is a small positive number. Here $H_{j}(\mathbf{x})$ is a part
of the CHF in interval $\Omega_{j}$, $\chi_{j}(t)$ is the indicator function
which takes value $1$ if $t\in\Omega_{j}$, and $0$ otherwise.

It is important that $H_{j}(\mathbf{x})$ does not depend on $t$ and is
constant in interval $\Omega_{j}$. The last condition will be necessary below
in order to deal with logarithms of the CHFs.

Let $g$ be a monotone function. Then there holds
\begin{equation}
g(H(t|\mathbf{x}))=\sum_{j=0}^{m}g(H_{j}(\mathbf{x}))\chi_{j}(t).
\end{equation}

The same expressions can be written for the CHF provided by the Cox model%
\begin{align}
H_{\text{Cox}}(t|\mathbf{x},\mathbf{b})  &  =H_{0}(t)\exp\left(
\mathbf{xb}^{\mathrm{T}}\right) \nonumber\\
&  =\sum_{j=0}^{m}\left[  H_{0j}\exp\left(  \mathbf{xb}^{\mathrm{T}}\right)
\right]  \chi_{j}(t),\ H_{0j}\geq\varepsilon.
\end{align}

The introduced condition $h_{j}\geq\varepsilon>0$ allows us to use logarithms.
Therefore, the distance between two CHFs is replaced with the distance between
logarithms of the corresponding CHFs for the optimization problem. Let
$\phi(t|\mathbf{x}_{k})$ and $\phi_{\text{Cox}}(t|\mathbf{x}_{k},\mathbf{b})$
be logarithms of $H(t|\mathbf{x}_{k})$ and $H_{\text{Cox}}(t|\mathbf{x}%
_{k},\mathbf{b})$, respectively. Here $\mathbf{x}_{k}$ is a generated point in
a local area around $\mathbf{x}$. The difference between functions
$\phi(t|\mathbf{x}_{k})$ and $\phi_{\text{Cox}}(t|\mathbf{x}_{k},\mathbf{b})$
can be written as follows:
\begin{align}
&  \phi(t|\mathbf{x}_{k})-\phi_{\text{Cox}}(t|\mathbf{x}_{k},\mathbf{b}%
)\nonumber\\
&  =\sum_{j=0}^{m}(\ln H_{j}(\mathbf{x}_{k}))\chi_{j}(t)-\sum_{j=0}^{m}\left(
\ln(H_{0j}\exp\left(  \mathbf{x}_{k}\mathbf{b}^{\mathrm{T}}\right)  )\right)
\chi_{j}(t)\nonumber\\
&  =\sum_{j=0}^{m}\left(  \ln H_{j}(\mathbf{x}_{k})-\ln H_{0j}-\mathbf{x}%
_{k}\mathbf{b}^{\mathrm{T}}\right)  \chi_{j}(t).
\end{align}

Generally, the $L_{p}$-norm can be used to write the distance between
$\phi(t|\mathbf{x}_{k})$ and $\phi_{\text{Cox}}(t|\mathbf{x}_{k},\mathbf{b})$.
However, we consider the $L_{\infty}$-norm or the Chebyshev distance because
it leads to a maximin optimization problem which is rather simple from
computational point of view and can be solved in a reasonable time. The
$L_{\infty}$-norm is a measure of the approximation quality, which is defined
as the maximum of the absolute values or the maximum deviation between the
function being approximated and the approximating function. Several authors
\cite{Marosevic-1996,Sim-Hartley-2006} pointed out that $L_{\infty}$
minimization is not robust to outliers, i.e., $L_{\infty}$ minimization may
fit the outliers and not the good data. However, this disadvantage can be
compensated by the introduce imprecision.

The distance between $\phi(t|\mathbf{x}_{k})$ and $\phi_{\text{Cox}%
}(t|\mathbf{x}_{k},\mathbf{b})$ is defined as
\begin{align}
D_{\infty,k}\left(  \phi,\phi_{\text{Cox}}\right)   &  =\left\Vert
\phi(t|\mathbf{x}_{k})-\phi_{\text{Cox}}(t|\mathbf{x}_{k},\mathbf{b}%
)\right\Vert _{\infty}\nonumber\\
&  =\max_{t\in\Omega}\left\vert \phi(t|\mathbf{x}_{k})-\phi_{\text{Cox}%
}(t|\mathbf{x}_{k},\mathbf{b})\right\vert .
\end{align}

Hence, the optimization problem for computing $\mathbf{b}$ is
\begin{equation}
\min_{\mathbf{b}}\left(  \sum_{k=1}^{N}w_{k}\cdot\max_{t\in\Omega}\left\vert
\phi(t|\mathbf{x}_{k})-\phi_{\text{Cox}}(t|\mathbf{x},\mathbf{b})\right\vert
\right)  . \label{KS_Inf_SurvLIME_50}%
\end{equation}

Let us introduce the optimization variables%
\begin{equation}
z_{k}=\max_{t\in\Omega}\left\vert \phi(t|\mathbf{x}_{k})-\phi_{\text{Cox}%
}(t|\mathbf{x}_{k},\mathbf{b})\right\vert ,\ k=1,...,N.
\end{equation}

They are restricted as follows:
\begin{equation}
z_{k}\geq\left\vert \phi(t|\mathbf{x}_{k})-\phi_{\text{Cox}}(t|\mathbf{x}%
_{k},\mathbf{b})\right\vert ,\ \forall t\in\Omega.
\end{equation}
Every above constraint can be represented as the following pair of
constraints:
\begin{equation}
z_{k}\geq\phi(t|\mathbf{x}_{k})-\phi_{\text{Cox}}(t|\mathbf{x}_{k}%
,\mathbf{b}),\ \forall t\in\Omega,
\end{equation}%
\begin{equation}
z_{k}\geq\phi_{\text{Cox}}(t|\mathbf{x}_{k},\mathbf{b})-\phi(t|\mathbf{x}%
_{k}),\ \forall t\in\Omega.
\end{equation}

Now the optimization problem (\ref{KS_Inf_SurvLIME_50}) is rewritten as
follows:
\begin{equation}
\min_{\mathbf{b}}\sum_{k=1}^{N}w_{k}z_{k},
\end{equation}
subject to
\begin{equation}
z_{k}\geq\sum_{j=0}^{m}\left(  \ln H_{j}(\mathbf{x}_{k})-\ln H_{0j}%
-\mathbf{x}_{k}\mathbf{b}^{\mathrm{T}}\right)  \chi_{j}(t),\ \forall
t\in\Omega,\ k=1,...,N,
\end{equation}%
\begin{equation}
z_{k}\geq\sum_{j=0}^{m}\left(  \ln H_{0j}+\mathbf{x}_{k}\mathbf{b}%
^{\mathrm{T}}-\ln H_{j}(\mathbf{x}_{k})\right)  \chi_{j}(t),\ \forall
t\in\Omega,\ k=1,...,N.
\end{equation}

Let us denote $\ln H_{j}(\mathbf{x}_{k})-\ln H_{0j}=\theta_{kj}$ for short.
Since the function $\theta_{kj}-\mathbf{x}_{k}\mathbf{b}^{\mathrm{T}}$ is $0$
for all $t\notin\Omega_{j}$, and it is constant for all $t\in\Omega_{j}$, then
the last constraints can be rewritten as
\begin{equation}
z_{k}\geq\theta_{kj}-\mathbf{x}_{k}\mathbf{b}^{\mathrm{T}},\ j=0,...,m,
\end{equation}%
\begin{equation}
z_{k}\geq\mathbf{x}_{k}\mathbf{b}^{\mathrm{T}}\mathbf{-}\theta_{kj}%
,\ j=0,...,m.
\end{equation}

The term $\mathbf{x}_{k}\mathbf{b}^{\mathrm{T}}$ does not depend on $j$.
Hence, the constraints can be reduced to the following simple constraints:%
\begin{equation}
z_{k}\geq Q_{k}-\mathbf{x}_{k}\mathbf{b}^{\mathrm{T}},\ k=1,...,N,
\label{KS_Inf_SurvLIME_37}%
\end{equation}%
\begin{equation}
z_{k}\geq\mathbf{x}_{k}\mathbf{b}^{\mathrm{T}}-R_{k},\ k=1,...,N.
\label{KS_Inf_SurvLIME_38}%
\end{equation}
where%
\begin{equation}
Q_{k}=\max_{j=0,...,m}\theta_{kj},\ R_{k}=\min_{j=0,...,m}\theta_{kj}.
\end{equation}

Finally, we get the linear optimization problem with $d+N$ optimization
variables ($z_{1},...,z_{N}$ and $\mathbf{b}$) and $2N$ constraints. It is of
the form:
\begin{equation}
\min_{\mathbf{b}}\sum_{k=1}^{N}w_{k}z_{k}, \label{KS_Inf_SurvLIME_36}%
\end{equation}
subject to (\ref{KS_Inf_SurvLIME_37})-(\ref{KS_Inf_SurvLIME_38}).

\section{Kolmogorov-Smirnov bounds for robustifying explanations}

Let us consider how to take into account the fact that the black-box model
provides unreliable and inaccurate results due to a restricted number of
training data or outliers. Let us return to the CHF $H(t|\mathbf{x})$
representation (\ref{KS_Inf_SurvLIME_20}) under condition $H_{j}%
(\mathbf{x})\geq\varepsilon>0$ considered above. We also assume that
$H_{j}(\mathbf{x})\leq H_{j+1}(\mathbf{x})$. Let us introduce the following
function:
\begin{align}
F(t|\mathbf{x})  &  =\frac{H(t|\mathbf{x})-\varepsilon}{\max_{t\in\Omega
}H(t|\mathbf{x})-\varepsilon}\nonumber\\
&  =\sum_{j=0}^{m}\frac{H_{j}(\mathbf{x})-\varepsilon}{H_{m}(\mathbf{x}%
)-\varepsilon}\chi_{j}(t)=\sum_{j=0}^{m}f_{j}(\mathbf{x})\chi_{j}(t).
\end{align}

The function $f_{j}(\mathbf{x})$ behaves like the empirical cumulative
distribution function $F(\mathbf{x})$ constructed on the basis of $n$
observations. This implies that KS bounds can be defined for this function
with a band of width $\pm d_{n,1-\gamma}$ as follows:
\begin{equation}
F^{L}(\mathbf{x})=\sum_{j=0}^{m}\max(f_{j}(\mathbf{x})-d_{n,1-\gamma}%
,0)\chi_{j}(t),
\end{equation}%
\begin{equation}
F^{U}(\mathbf{x})=\sum_{j=0}^{m}\min(f_{j}(\mathbf{x})+d_{n,1-\gamma}%
,1)\chi_{j}(t).
\end{equation}

By returning to the CHF, we get bounds $H^{L}(\mathbf{x})$ and $H^{U}%
(\mathbf{x})$ for $H(t|\mathbf{x})$:
\begin{equation}
H^{L}(t|\mathbf{x})=\sum_{j=0}^{m}\max(H_{j}(\mathbf{x})-\Delta_{1-\gamma
},0)\chi_{j}(t),
\end{equation}%
\begin{equation}
H^{U}(t|\mathbf{x})=\sum_{j=0}^{m}\min(H_{j}(\mathbf{x})+\Delta_{1-\gamma
},1)\chi_{j}(t).
\end{equation}

Here the value of $\Delta_{1-\gamma}$ is derived as follows. Let us take
values $j$ such that $H_{j}(\mathbf{x})-\Delta_{1-\gamma}\geq0$. Then there
holds for the lower bound
\begin{equation}
f_{j}(\mathbf{x})-d_{n,1-\gamma}=\frac{H_{j}(\mathbf{x})-\varepsilon}%
{H_{m}(\mathbf{x})-\varepsilon}-d_{n,1-\gamma}.
\end{equation}
Hence, we write
\begin{equation}
\left(  H_{m}(\mathbf{x})-\varepsilon\right)  f_{j}(\mathbf{x})-\left(
H_{m}(\mathbf{x})-\varepsilon\right)  d_{n,1-\gamma}=\left(  H_{j}%
(\mathbf{x})-\varepsilon\right)  -\left(  H_{m}(\mathbf{x})-\varepsilon
\right)  d_{n,1-\gamma}.
\end{equation}

It follows from the above that
\begin{align}
&  \left\{  \left(  H_{m}(\mathbf{x})-\varepsilon\right)  f_{j}(\mathbf{x}%
)+\varepsilon\right\}  -\left(  H_{m}(\mathbf{x})-\varepsilon\right)
d_{n,1-\gamma}\nonumber\\
&  =H_{j}(\mathbf{x})-\left(  H_{m}(\mathbf{x})-\varepsilon\right)
d_{n,1-\gamma}=H_{j}(\mathbf{x})-\Delta_{1-\gamma}.
\end{align}
Hence, there holds $\Delta_{1-\gamma}=\left(  H_{m}(\mathbf{x})-\varepsilon
\right)  d_{n,1-\gamma}$.

In the same way, we can consider the upper bound.

It should be noted that the obtained bounds cannot be regarded as true KS
bounds for an empirical cumulative distribution function. Therefore, we do not
define these bounds as a function of the number of observations $n$. The
bounds or confidence $\gamma$ should be viewed as a tuning parameter of the
machine learning algorithm. At the same time, we can always imagine a
\textquotedblleft normalized\textquotedblright\ non-decreasing step-wise
function as an empirical cumulative distribution function of some random
variable. The corresponding KS bounds in this case provide confidence bounds
for a set of possible probability distributions. Therefore, the inverse
transformation of the bounds presented above provides some type of confidence
intervals for the CHFs under condition that the CHF is bounded above.

The logarithmic function is monotone. Therefore, we can write bounds for the
function $\phi_{kj}=\ln H_{j}(\mathbf{x}_{k})$ as%
\begin{equation}
\phi_{k,j}^{L}\leq\phi_{k,j}\leq\phi_{k,j}^{U}, \label{KS_Inf_SurvLIME_46}%
\end{equation}
where
\begin{equation}
\phi_{k,j}^{L}=\ln\left(  \max(H_{j}(\mathbf{x}_{k})-\Delta_{1-\gamma
},0)\right)  ,
\end{equation}%
\begin{equation}
\phi_{k,j}^{U}=\ln\left(  \min(H_{j}(\mathbf{x})+\Delta_{1-\gamma},1)\right)
.
\end{equation}

We have to point out that $\phi_{kj}$ are elements of a non-decreasing
function (logarithm of the CHF). Therefore, we have to extend constraints for
$\phi_{kj}$ by the following constraints:
\begin{equation}
\phi_{k,j}\leq\phi_{k,j+1},\ j=0,...,m-1. \label{KS_Inf_SurvLIME_47}%
\end{equation}

Let us replace the values $\phi_{k,j}$ with values $\theta_{k,j}=\phi
_{k,j}-\ln H_{0j}$ for simplicity purposes. Then constraints
(\ref{KS_Inf_SurvLIME_46}) and (\ref{KS_Inf_SurvLIME_47}) can be rewritten as
follows:
\begin{equation}
\theta_{k,j}^{L}\leq\theta_{k,j}\leq\theta_{k,j}^{U},\ k=1,...,N,\ j=0,...,m,
\label{KS_Inf_SurvLIME_48}%
\end{equation}%
\begin{equation}
\theta_{k,j}\leq\theta_{k,j+1}+\ln(H_{0j+1}/H_{0j}),\ k=1,...,N,\ j=0,...,m-1,
\label{KS_Inf_SurvLIME_49}%
\end{equation}

Constraints (\ref{KS_Inf_SurvLIME_48}) and (\ref{KS_Inf_SurvLIME_49}) produce
a set $\mathcal{F}_{k}$ of vectors $\mathbf{\theta}_{k}=(\theta_{k0}%
,...,\theta_{km})$ for every $k$.

It is proposed to robustify the explanation algorithm by replacing the precise
values of $\theta_{kj}$ obtained as an output of the black-box model with
their intervals $[\theta_{kj}^{L},\theta_{kj}^{U}]$. One of the well-known
ways for dealing with the interval-valued expected risk is to use the maximin
(pessimistic or robust) strategy for which a vector $\mathbf{\theta}_{k}$ is
selected from the set $\mathcal{F}_{k}$ such that the loss function
$L(\mathbf{b})$ achieves its largest value or its upper bound for fixed values
of $\mathbf{b}$. In other words, we use the upper bound of $L(\mathbf{b})$ for
computing optimal $\mathbf{b}$. Since the maximin strategy provides the
largest value of the expected loss, then it can be interpreted as an insurance
against the worst case because it aims at minimizing the expected loss in the
least favorable case \cite{Robert94}. In sum, the following maximin
optimization problem can be written:
\begin{equation}
\max_{\mathbf{\theta}_{k}}\min_{\mathbf{b,}z_{k}}\sum_{k=1}^{N}w_{k}z_{k},
\label{KS_Inf_SurvLIME_52}%
\end{equation}
subject to (\ref{KS_Inf_SurvLIME_37}) and (\ref{KS_Inf_SurvLIME_38}),
$\mathbf{\theta}_{k}\in\mathcal{F}_{k}$, $k=1,...,N$.

First, we consider the minimization problem and write the corresponding dual
one. Introduce non-negative vectors of variables $\mathbf{a}$ and $\mathbf{c}$
such that $\mathbf{b=a}-\mathbf{c}$. Then we get the problem with non-negative
variables%
\begin{equation}
\min_{\mathbf{b,}z_{k}}\sum_{k=1}^{N}w_{k}z_{k},
\end{equation}
subject to $z_{k}\geq0$, $k=1,...,N$, $a_{i}\geq0$, $b_{i}\geq0$, $i=1,...,d$,
and
\begin{equation}
z_{k}+\mathbf{x}_{k}\mathbf{a}^{\mathrm{T}}-\mathbf{x}_{k}\mathbf{c}%
^{\mathrm{T}}\geq Q_{k},\ k=1,...,N,
\end{equation}%
\begin{equation}
z_{k}-\mathbf{x}_{k}\mathbf{a}^{\mathrm{T}}+\mathbf{x}_{k}\mathbf{c}%
^{\mathrm{T}}\geq-R_{k},\ k=1,...,N.
\end{equation}
The dual problem is of the form:
\begin{equation}
\max_{\alpha,\beta}\sum_{k=1}^{N}\left(  Q_{k}\alpha_{k}-R_{k}\beta
_{k}\right)  ,
\end{equation}
subject to $\alpha_{k}\geq0$, $\beta_{k}\geq0$, and
\begin{equation}
\alpha_{k}+\beta_{k}\leq w_{k},\ k=1,...,N, \label{KS_Inf_SurvLIME_56}%
\end{equation}%
\begin{equation}
\sum_{k=1}^{N}x_{k}^{(i)}\left(  \alpha_{k}-\beta_{k}\right)  =0,\ i=1,...,d.
\label{KS_Inf_SurvLIME_57}%
\end{equation}

Now the maximin problem (\ref{KS_Inf_SurvLIME_52}) can be rewritten as the
following maximization problem:
\begin{equation}
\max_{\mathbf{\theta}_{k},\alpha,\beta}\sum_{k=1}^{N}\left(  Q_{k}\alpha
_{k}-R_{k}\beta_{k}\right)  , \label{KS_Inf_SurvLIME_60}%
\end{equation}
subject to $\alpha_{k}\geq0$, $\beta_{k}\geq0$, (\ref{KS_Inf_SurvLIME_56}),
(\ref{KS_Inf_SurvLIME_57}), (\ref{KS_Inf_SurvLIME_48}),
(\ref{KS_Inf_SurvLIME_49}).

All constraints in (\ref{KS_Inf_SurvLIME_60}) are linear. Moreover, it is
interesting to note that constraints for $\theta_{k,j}$ differ from
constraints for $\alpha_{k}$ and $\beta_{k}$.

Let us consider the objective function in detail. First, if we fix all
variables $\alpha_{k}$, $\beta_{k}$, then it can be seen that optimal values
of $Q_{k}$ and $R_{k}$ do not depend on optimal values of $Q_{l}$ and $R_{l}$,
$l\neq k$, respectively. Indeed, constraints (\ref{KS_Inf_SurvLIME_48}) and
(\ref{KS_Inf_SurvLIME_49}) are different for every $k$ and do not intersect
each other. Second, it is obvious that variable $Q_{k}$ has to be as large as
possible, and variable $R_{k}$ has to be as small as possible. Let us prove
that their optimal values do not depend on variables $\alpha_{k}$ and
$\beta_{k}$. Indeed, it follows from the definition of $Q_{k}$ and $R_{k}$
that, by solving the linear optimization problems with constraints
(\ref{KS_Inf_SurvLIME_48}) and (\ref{KS_Inf_SurvLIME_49}), we find the largest
value $\theta_{kj}=\theta_{kq}$ corresponding to some value $j=q$, i.e.,
$Q_{k}=\theta_{kq}$. The obtained value $\theta_{kq}$ belongs to the set
$\mathcal{F}_{k}$, i.e., it belongs to the feasible region defined by
constraints (\ref{KS_Inf_SurvLIME_48}) and (\ref{KS_Inf_SurvLIME_49}). In the
same way, we find the smallest value $\theta_{kj}=\theta_{kr}$ corresponding
to some value $j=r$, i.e., $R_{k}=\theta_{kr}$. The obtained value
$\theta_{kr}$ belongs to the set $\mathcal{F}_{k}$. The values $q$ and $r$ are
different if there is at least one point which differ two functions $\ln
H_{j}(\mathbf{x}_{k})$ and $\ln H_{0j}$, i.e., the function $\theta_{k,j}$ has
a non-zero value at least at one $j$. If $q$ and $r$ are obtained identical,
then $Q_{k}=R_{k}=0$ and the corresponding term in the sum
(\ref{KS_Inf_SurvLIME_60}) is not used. As a results, we can separately find
values of $Q_{k}$ and $R_{k}$ for every $k=1,...,N$, as follows:
\begin{equation}
Q_{k}=\max_{j=0,...,m}\left(  \max\theta_{kj}\right)  ,\ R_{k}=\min
_{j=0,...,m}\left(  \min\theta_{kj}\right)  ,
\end{equation}
subject to (\ref{KS_Inf_SurvLIME_48}) and (\ref{KS_Inf_SurvLIME_49}).

Let us prove that optimal values of $Q_{k}$ and $R_{k}$ are $\max
_{j=0,...,m}\theta_{k,j}^{U}$ and $\min_{j=0,...,m}\theta_{k,j}^{L}$,
respectively. Assume that we have only conditions (\ref{KS_Inf_SurvLIME_48}).
Then it is obvious that the optimal value of $\theta_{kj}$ for computing
$Q_{k}$ is $\theta_{k,j}^{U}$. It is easy to show that this optimal value
satisfies constraints (\ref{KS_Inf_SurvLIME_49}). Let us return to the same
constraint in the form of $\phi_{k,j}$, i.e., (\ref{KS_Inf_SurvLIME_47}),
where $\phi_{k,j}^{U}$ corresponds to $\theta_{k,j}^{U}$. One can see that
$\phi_{k,j}^{U}\leq\phi_{k,j+1}^{U}$ because the function $\phi_{k,j}$ is
increasing with $j$. This implies that we can always find $\phi_{k,j+1}$ which
is larger than $\phi_{k,j}^{U}$, for example, $\phi_{k,j+1}^{U}$. The same can
be said about $\theta_{k,j}^{U}$ and $\theta_{k,j+1}^{U}$ because
$\theta_{k,j}^{U}=\phi_{k,j}^{U}-\ln H_{0,j}$ and $\theta_{k,j+1}^{U}%
=\phi_{k,j+1}^{U}-\ln H_{0,j+1}$. In sum, we have the optimal value
$\theta_{k,j}^{U}$ of $\theta_{k,j}$. Hence, the optimal value of $Q_{k}$ is
$\max_{j=0,...,m}\theta_{k,j}^{U}$. A similar proof can be given for $R_{k}$.

By using the obtained results, we can return to the primal optimization
problem (\ref{KS_Inf_SurvLIME_36})-(\ref{KS_Inf_SurvLIME_38}):
\begin{equation}
\min_{\mathbf{b}}\sum_{k=1}^{N}w_{k}z_{k},
\end{equation}
subject to%
\begin{equation}
z_{k}+\mathbf{x}_{k}\mathbf{b}^{\mathrm{T}}\geq\max_{j=0,...,m}\theta
_{k,j}^{U},\ k=1,...,N, \label{KS_Inf_SurvLIME_68}%
\end{equation}%
\begin{equation}
z_{k}-\mathbf{x}_{k}\mathbf{b}^{\mathrm{T}}\geq-\min_{j=0,...,m}\theta
_{k,j}^{L},\ k=1,...,N. \label{KS_Inf_SurvLIME_69}%
\end{equation}

The linear optimization problem is computationally simple, but it may have the
sparse solution because an optimal $\mathbf{b}$ can be found among extreme
points of the feasible set, In order to restrict the space of admissible
solutions, we add the standard Tikhonov regularization term which can be
regarded as a constraint which enforces uniqueness by penalizing functions
with wild oscillation. As a result, we get the following objective function
\begin{equation}
\min_{\mathbf{b}}\sum_{k=1}^{N}w_{k}z_{k}+\lambda\left\Vert \mathbf{b}%
\right\Vert ^{2}, \label{KS_Inf_SurvLIME_70}%
\end{equation}
where $\lambda$ is a hyper-parameter which controls the strength of the regularization.

Finally, we get a way to implement the robust explanation. Objective function
(\ref{KS_Inf_SurvLIME_70}) and constraints (\ref{KS_Inf_SurvLIME_68}%
)-(\ref{KS_Inf_SurvLIME_69}) compose the quadratic optimization problem whose
solution is a standard task.

\section{Numerical experiments}

To perform numerical experiments, we use the following general scheme.

1. The black-box Cox model and the RSF are trained on synthetic or real
survival data. The outputs of the trained models in the testing phase are CHFs
and SFs.

2. In order to analyze and to compare results of experiments for different
initial data, we consider two measures: the Root Square Error (RSE) and the
Mean Root Square Error (MRSE). These measures are defined as
\begin{equation}
RSE(H_{\text{model}},H_{\text{approx}})=\sqrt{\frac{1}{m+1}\sum_{j=0}%
^{m}\left(  H_{\text{model}}^{(j)}(\mathbf{x})-H_{\text{approx}}%
^{(j)}(\mathbf{x})\right)  ^{2}},
\end{equation}%
\begin{equation}
MRSE(H_{\text{model}},H_{\text{approx}})=\frac{1}{n}\sum_{i=1}^{n}RSE\left(
H_{\text{model}}(t|\mathbf{x}_{i}),H_{\text{approx}}(t|\mathbf{x}_{i})\right)
.
\end{equation}

Here $H_{\text{model}}^{(j)}(t|\mathbf{x})$ and $H_{\text{approx}}%
^{(j)}(t|\mathbf{x})$ are two compared CHFs under condition that $t\in
\Omega_{j}$, $j=0,...,m$, i.e., $H_{\text{model}}^{(j)}(\mathbf{x}%
)=H_{\text{model}}(t|\mathbf{x})$ and $H_{\text{approx}}^{(j)}(\mathbf{x}%
)=H_{\text{approx}}(t|\mathbf{x})$ by $t\in\Omega_{j}$. In other words, we
consider differences between values of CHFs at intervals $\Omega_{j}$. The CHF
$H_{\text{model}}$ is obtained by testing the black-box model (the Cox model
or the RSF), and the CHF $H_{\text{approx}}$ is computed from the Cox
approximation by substituting into the model optimal values of coefficients
$\mathbf{b}$. $RSE$ can be regarded as a normalized distance between two
corresponding discrete functions, i.e. it characterizes the difference between
the CHFs $H_{\text{model}}$ and $H_{\text{approx}}$. The second measure $MRSE$
is the mean of $RSE$s defined for $n$ testing points.

3. In order to study the proposed explanation algorithm by means of synthetic
data, we generate random survival times to events by using the Cox model
estimates. This generation allows us to compare initial data for generating
every points and results of SurvLIME-KS.

4. We investigate the approximating Cox model by changing the hyper-parameter
$\lambda$ of the regularization and the confidence probability $1-\gamma$ of
KS bounds. The parameter $\lambda$ takes values from interval $\Lambda
_{\text{Cox}}=[10^{-1},10^{4}]$ for the black-box Cox model and from interval
$\Lambda_{\text{RSF}}=[10^{-1},10^{6}]$ for the RSF. The probability $\gamma$
is taken from the set $\Gamma=\{0.005,0.01,0.05,0.1,1\}$. The case $\gamma=1$
corresponds to the lack of bounds.

5. In order to compare different cases, we train every black-box model (the
Cox model and the RSF) on a large dataset consisting of $200$ examples, on a
small dataset consisting of $20$ examples which are randomly selected from the
large dataset. The corresponding models will be denoted as $M_{\text{large}}$
and $M_{\text{small}}$. We compute three measures: $E_{1}=RSE\left(
H_{M_{\text{large}}},H_{\text{approx1}}\right)  $, $E_{2}=RSE\left(
H_{M_{\text{small}}},H_{\text{approx2}}\right)  $, $E_{3}=RSE\left(
H_{M_{\text{small}}},H_{\text{approx3}}\right)  $, where $H_{\text{approx1}}$
and $H_{\text{approx2}}$ are CHFs provided by the approximating Cox model
without KS bounds for black-box models (the Cox model and the RSF)
$M_{\text{large}}$ and $M_{\text{small}}$, respectively; $H_{\text{approx3}}$
is the CHF provided by the approximating Cox model with KS bounds for
$M_{\text{small}}$. In other words, we perform experiments with the large
dataset and then cut it to get the small dataset in order to study whether the
use of KS bounds provides better results in comparison with the models trained
without KS bounds.

\subsection{Synthetic data}

\subsubsection{Initial parameters of numerical experiments with synthetic
data}

Synthetic training and testing sets are composed as follows. Random survival
times to events are generated by using the Cox model estimates. For performing
numerical experiments, $N=1000$ covariate vectors $\mathbf{x}\in\mathbb{R}%
^{d}$ are randomly generated from the uniform distribution in the $d$-sphere
with predefined radius $R=8$. Here $d=5$. The center of the sphere is
$p=(0,0,0,0,0)$. There are several methods for the uniform sampling of points
$\mathbf{x}$ in the $d$-sphere with the unit radius $R=1$, for example,
\cite{Barthe-etal-2005,Harman-Lacko-2010}. Then every generated point is
multiplied by $R=8$.

In order to generate random survival times by using the Cox model estimates,
we apply results obtained by Bender et al. \cite{Bender-etal-2005} for
survival time data for the Cox model with Weibull distributed survival times.
The Weibull distribution with the scale $\lambda_{0}=10^{-5}$ and shape $v=2$
parameters is used to generate appropriate survival times because this
distribution shares the assumption of proportional hazards with the Cox
regression model \cite{Bender-etal-2005}. For experiments, we take the vector
$\mathbf{b}_{\text{true}}=(-0.25,10^{-6},-0.1,0.35,10^{-6})$, which has two
almost zero-valued elements ($10^{-6}$) and three \textquotedblleft
large\textquotedblright\ elements ($-0.25$, $-0.1$, $0.3$) which will
correspond to important features. Random survival times $T$ are generated in
accordance with \cite{Bender-etal-2005} by using parameters $\lambda_{0}$,
$v$, $\mathbf{b}_{\text{true}}$ as follows:%
\begin{equation}
T=\left(  \frac{-\ln(U)}{\lambda_{0}\exp\left(  \mathbf{xb}_{\text{true}%
}^{\mathrm{T}}\right)  }\right)  ^{1/v}, \label{KS_Inf_SurvLIME_84}%
\end{equation}
where $U$ is the random variable uniformly distributed in interval $[0,1]$.

Generated values $T_{i}$ are restricted by the condition: if $T_{i}>2000$,
then $T_{i}$ is replaced with value $2000$. The event indicator $\delta_{i}$
is generated from the binomial distribution with probabilities $\Pr
\{\delta_{i}=1\}=0.9$, $\Pr\{\delta_{i}=0\}=0.1$.

According to the proposed algorithm, $N$ nearest points $\mathbf{x}_{k}$ are
generated in a local area around point $\mathbf{x}$ as its perturbations.
These points are uniformly generated in the $d$-sphere with some predefined
radius $r=0.1$ and with the center at point $\mathbf{x}$. The weights to every
point $\mathbf{x}_{k}$ is assigned as follows:
\begin{equation}
w_{k}=1-\left(  r^{-1}\cdot\left\Vert \mathbf{x}-\mathbf{x}_{k}\right\Vert
_{2}\right)  ^{1/2}.
\end{equation}

\subsubsection{The black-box Cox model}

\begin{figure}
[ptb]
\begin{center}
\includegraphics[
%%=6.000100in,
%%=9.999800in,
height=2.4491in,
width=4.075in
]%
{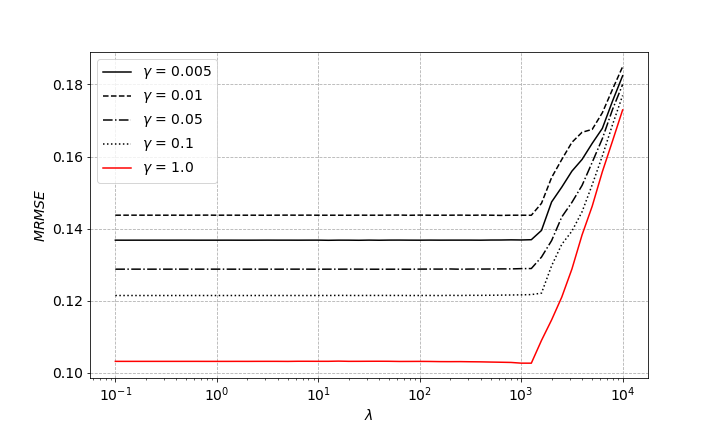}%
\caption{The MRSE as a function of hyper-parameter $\lambda$ and probability
$\gamma$ for the black-box Cox model}%
\label{f:cox_dep_regularization}%
\end{center}
\end{figure}

The first part of numerical experiments is performed with the black-box Cox
model. We study how the MRSE depends on the hyper-parameter $\lambda$ of the
regularization and the probability $\gamma$, which defines the interval of
CHFs when KS bounds are used. The corresponding results are shown in Fig.
\ref{f:cox_dep_regularization}. It can be seen from Fig.
\ref{f:cox_dep_regularization} that the smallest value of the MRSE is achieved
for $\gamma=1$. This implies that the lack of bounds leads to the best
results. The same can be seen from Table \ref{t:cox_synth_KS}, where the RSE
measures under different experiment conditions are shown for $10$ training
examples. The measure $E_{3}$ is obtained under condition $\gamma=0.1$. The
best results for the small dataset corresponding to the use of KS bounds is
shown in bold. We show in bold only these results because our aim is to study
how KS bounds impact on the explanation. It can be seen from Table
\ref{t:cox_synth_KS} that most examples ($7$ from $10$) show inferior results
for the KS bounds. This fact can be explained as follows. First of all,
training examples are generated from the same distribution. There are no
outliers in the dataset. Second, the black-box model (the Cox model) coincides
with the approximating model. At that, we do not approximate training data. We
approximate results of the black-box model. Therefore, the introduced
imprecision for the approximating Cox model by means of KS bounds makes the
approximation worse.%

%TCIMACRO{\FRAME{ftbpFU}{4.075in}{2.4491in}{0pt}{\Qcb{The MRSE as a function of
%hyper-parameter $\lambda$ and probability $\gamma$ for the black-box Cox
%model}}{\Qlb{f:cox_dep_regularization}}{cox_dep_regularization.png}%
%width 4.075in;  height 2.4491in;  depth 0pt;  original-width 9.9998in;
%original-height 6.0001in;  cropleft "0";  croptop "1";  cropright "1";
%filename 'cox_dep_regularization.png';file-properties "XNPEU";}} }%
%BeginExpansion
%EndExpansion
%

%TCIMACRO{\TeXButton{B}{\begin{table}[tbp] \centering}}%
%BeginExpansion
\begin{table}[tbp] \centering
%EndExpansion
\caption{RSE and MRSE for three cases of using KS bounds with the large and small datasets for training the black-box Cox model}%
\begin{tabular}
[c]{cccc}\hline
& \multicolumn{3}{c}{datasets}\\\hline
& large & \multicolumn{2}{c}{small}\\\hline
examples & $E_{1}$ & $E_{2}$ & $E_{3}$\\\hline
0 & $0.081$ & $0.090$ & $0.171$\\\hline
1 & $0.082$ & $0.084$ & $0.166$\\\hline
2 & $0.015$ & $0.068$ & $0.128$\\\hline
3 & $0.083$ & $0.101$ & $\mathbf{0.044}$\\\hline
4 & $0.082$ & $0.076$ & $\mathbf{0.038}$\\\hline
5 & $0.086$ & $0.092$ & $\mathbf{0.066}$\\\hline
6 & $0.073$ & $0.089$ & $0.181$\\\hline
7 & $0.025$ & $0.048$ & $0.053$\\\hline
8 & $0.079$ & $0.043$ & $0.057$\\\hline
9 & $0.080$ & $0.070$ & $0.116$\\\hline
$MRSE$ & $0.069$ & $0.076$ & $0.102$\\\hline
\end{tabular}
\label{t:cox_synth_KS}%
%TCIMACRO{\TeXButton{E}{\end{table}}}%
%BeginExpansion
\end{table}%
%EndExpansion

\begin{figure}
[ptb]
\begin{center}
\includegraphics[
%%=9.999800in,
%%=19.999599in,
height=2.3929in,
width=4.7746in
]%
{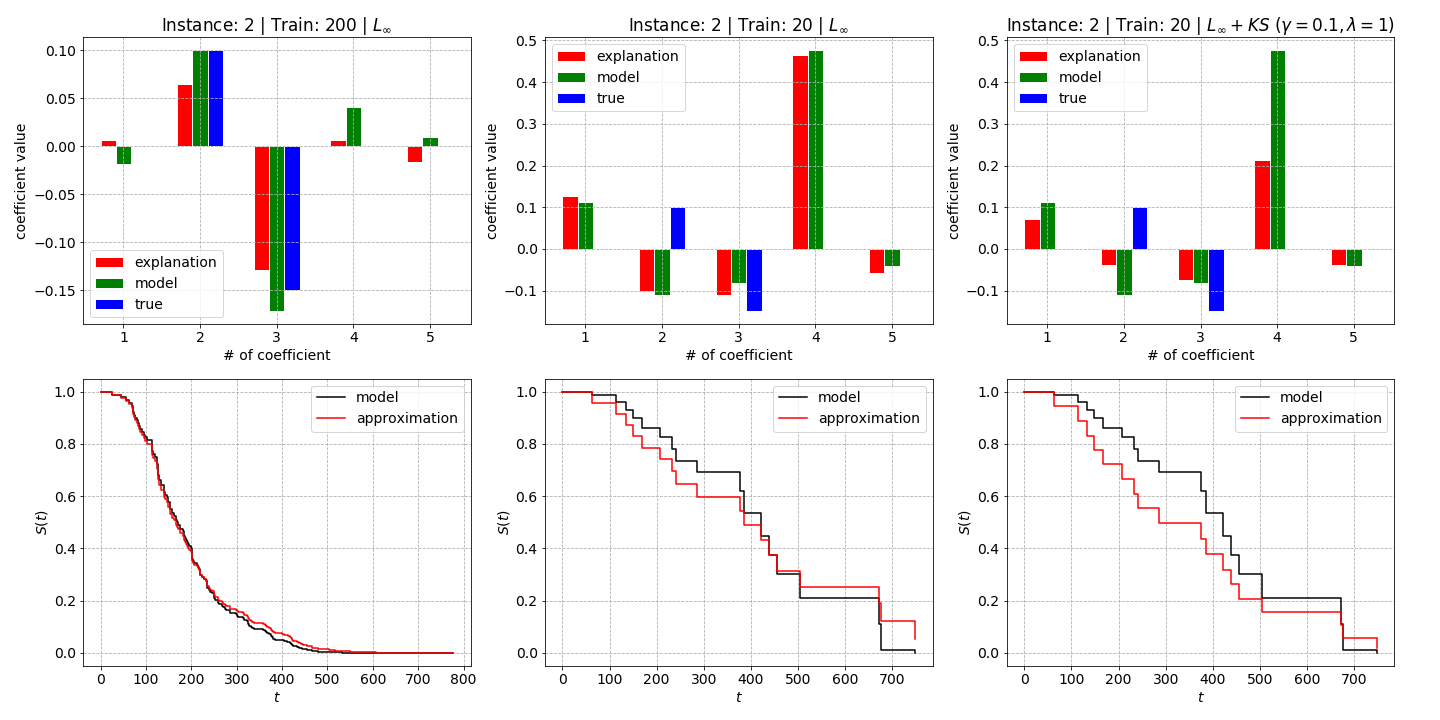}%
\caption{An illustration of the approximation results under three conditions
of experiments for the black-box Cox model by considering a \textquotedblleft
bad\textquotedblright\ example}%
\label{fig:cox_instance_2}%
\end{center}
\end{figure}

\begin{figure}
[ptb]
\begin{center}
\includegraphics[
%%=9.999800in,
%%=19.999599in,
height=2.38in,
width=4.7487in
]%
{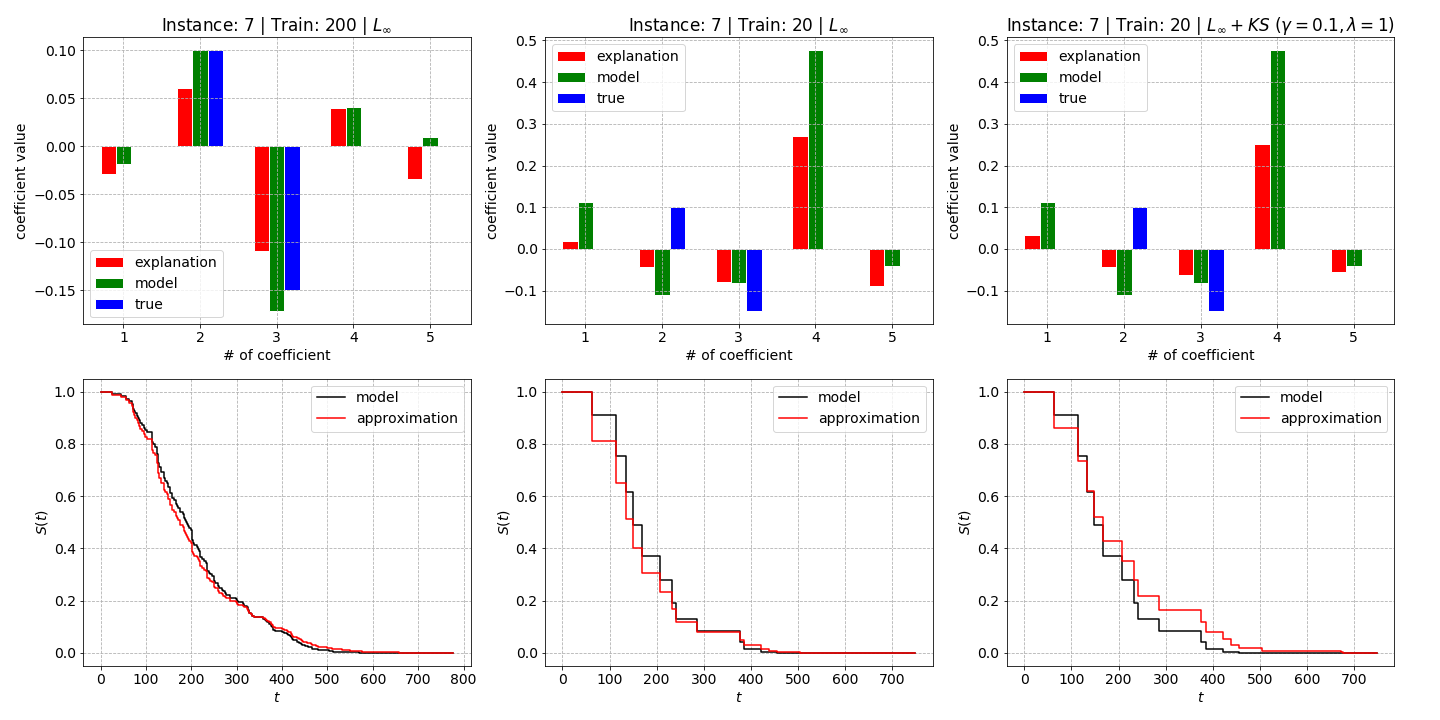}%
\caption{An illustration of the approximation results under three conditions
of experiments for the black-box Cox model by considering a \textquotedblleft
bad\textquotedblright\ example}%
\label{fig:cox_instance_7}%
\end{center}
\end{figure}

\begin{figure}
[ptb]
\begin{center}
\includegraphics[
%%=9.999800in,
%%=19.999599in,
height=2.38in,
width=4.7496in
]%
{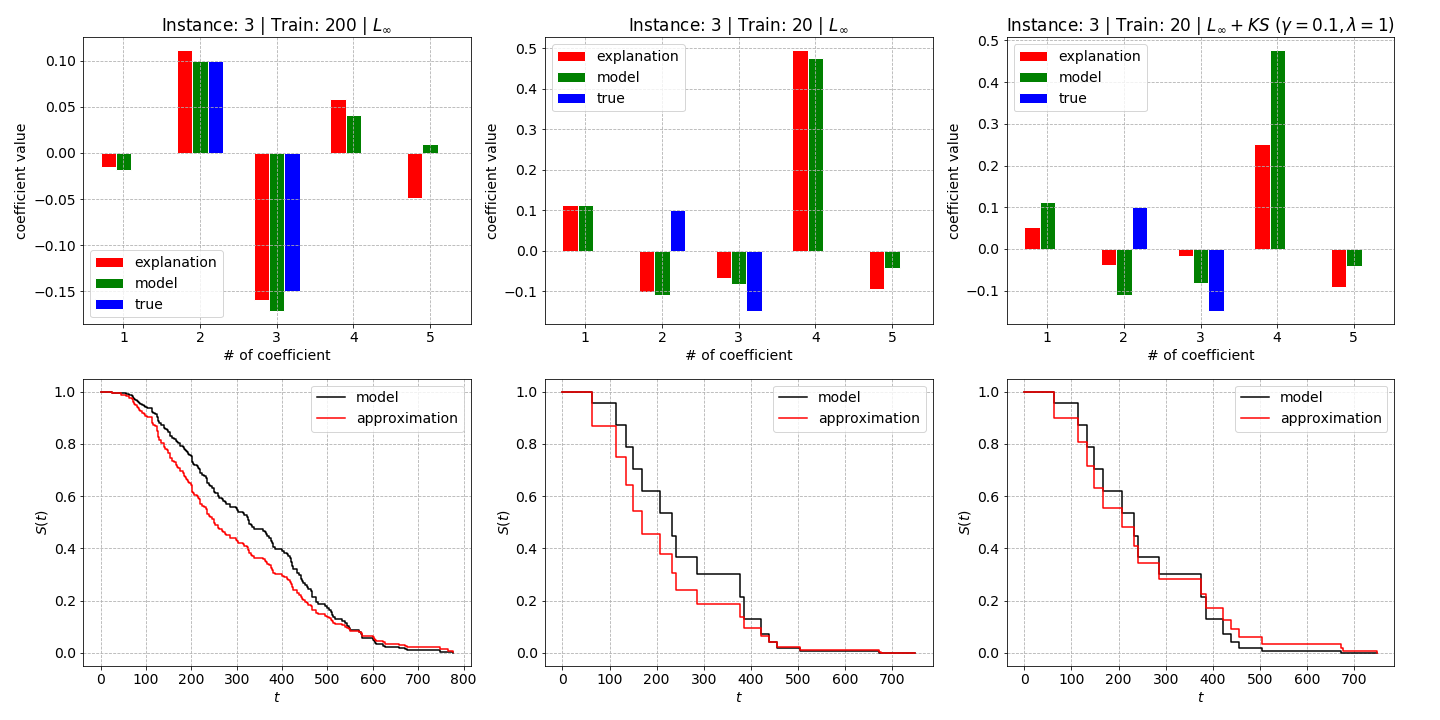}%
\caption{An illustration of the approximation results under three conditions
of experiments for the black-box Cox model by considering a \textquotedblleft
good\textquotedblright\ example}%
\label{fig:cox_instance_3}%
\end{center}
\end{figure}

The corresponding results are illustrated in Figs. \ref{fig:cox_instance_2}%
-\ref{fig:cox_instance_3}. In particular, Fig. \ref{fig:cox_instance_2} shows
three considered cases of experiments (every column of pictures). The first
row illustrates the coefficients $\mathbf{b}$ of important features for the
three conditions, where \textquotedblleft model\textquotedblright\ or
$\mathbf{b}_{\text{model}}$ are coefficients of the Cox model which is used as
the black-box model; \textquotedblleft true\textquotedblright\ or
$\mathbf{b}_{\text{true}}$ are coefficients used for training data generation
(see (\ref{KS_Inf_SurvLIME_84})); \textquotedblleft
explanation\textquotedblright\ or $\mathbf{b}_{\text{expl}}$ are explaining
coefficients obtained by using the proposed algorithm. One can see from Fig.
\ref{fig:cox_instance_2} that coefficients $\mathbf{b}_{\text{expl}}$ are
close to $\mathbf{b}_{\text{model}}$ and to $\mathbf{b}_{\text{true}}$ at the
first picture because the black-box Cox model is trained on the large dataset,
and this trained model is of the high quality. A different relationship is
observed in the second and the third pictures of the first row, which
correspond to condition of small data. At the same time, it can be seen from
these pictures that the explanation $\mathbf{b}_{\text{expl}}$ almost
coincides with the coefficients of the black-box Cox model $\mathbf{b}%
_{\text{model}}$. This is due to the fact that we explain the model results,
i.e., results obtained by the black-box Cox model, but not the training
dataset. The second row of pictures in Fig. \ref{fig:cox_instance_2}
illustrates how the SFs obtained by means of the black-box Cox model
(\textquotedblleft model\textquotedblright) and by means of the approximating
Cox model (\textquotedblleft approximation\textquotedblright) are close to
each other. It can be seen from these pictures that the corresponding SFs
under condition of large data are very close to each other. Moreover, we can
clearly see that the lack of imprecision for small data provides better
results than the use of KS bounds. Fig. \ref{fig:cox_instance_2} is an example
of a \textquotedblleft bad\textquotedblright\ case when the introduction of KS
bounds leads to the worse approximation and explanation. A similar example is
illustrated in Fig. \ref{fig:cox_instance_7}. In contrast to the examples in
Figs. \ref{fig:cox_instance_2} and \ref{fig:cox_instance_7}, Fig.
\ref{fig:cox_instance_3} illustrates a \textquotedblleft
good\textquotedblright\ case when the introduction of KS bounds leads to the
better approximation and explanation.%

%TCIMACRO{\FRAME{ftbpFU}{4.7746in}{2.3929in}{0pt}{\Qcb{An illustration of the
%approximation results under three conditions of experiments for the black-box
%Cox model by considering a \textquotedblleft bad\textquotedblright\ example}%
%}{\Qlb{fig:cox_instance_2}}{cox_instance_2.png}%
%width 4.7746in;  height 2.3929in;  depth 0pt;  original-width 19.9996in;
%original-height 9.9998in;  cropleft "0";  croptop "1";  cropright "1";
%cropbottom "0";  filename 'cox_instance_2.png';file-properties "XNPEU";}} }%
%BeginExpansion
%EndExpansion
%

%TCIMACRO{\FRAME{ftbpFU}{4.7487in}{2.38in}{0pt}{\Qcb{An illustration of the
%approximation results under three conditions of experiments for the black-box
%Cox model by considering a \textquotedblleft bad\textquotedblright\ example}%
%}{\Qlb{fig:cox_instance_7}}{cox_instance_7.png}%
%width 4.7487in;  height 2.38in;  depth 0pt;  original-width 19.9996in;
%original-height 9.9998in;  cropleft "0";  croptop "1";  cropright "1";
%cropbottom "0";  filename 'cox_instance_7.png';file-properties "XNPEU";}} }%
%BeginExpansion
%EndExpansion
%

%TCIMACRO{\FRAME{ftbpFU}{4.7496in}{2.38in}{0pt}{\Qcb{An illustration of the
%approximation results under three conditions of experiments for the black-box
%Cox model by considering a \textquotedblleft good\textquotedblright\ example}%
%}{\Qlb{fig:cox_instance_3}}{cox_instance_3.png}%
%width 4.7496in;  height 2.38in;  depth 0pt;  original-width 19.9996in;
%original-height 9.9998in;  cropleft "0";  croptop "1";  cropright "1";
%cropbottom "0";  filename 'cox_instance_3.png';file-properties "XNPEU";}} }%
%BeginExpansion
%EndExpansion

\subsubsection{The RSF}

\begin{figure}
[ptb]
\begin{center}
\includegraphics[
%%=6.000100in,
%%=9.999800in,
height=2.4915in,
width=4.1451in
]%
{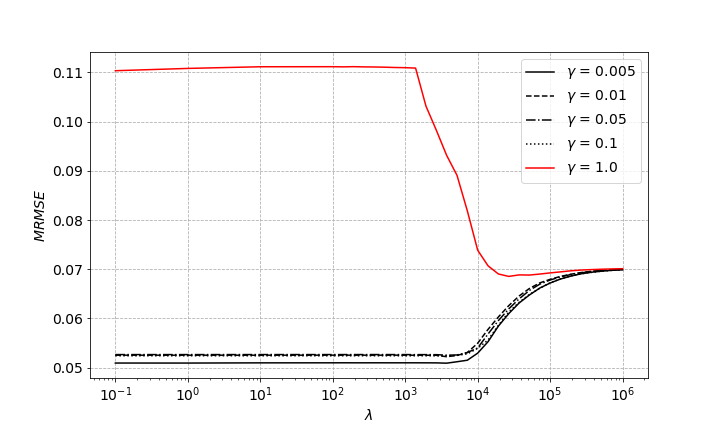}%
\caption{The MRSE as a function of hyper-parameter $\lambda$ and probability
$\gamma$ for the RSF}%
\label{f:rsf_dep_regularization}%
\end{center}
\end{figure}

The second part of numerical experiments is performed with the RSF as a
black-box model. We again study how the MRSE depends on the hyper-parameter
$\lambda$ of the regularization and the probability $\gamma$. The
corresponding results are shown in Fig. \ref{f:rsf_dep_regularization}. We can
observe quite different results. It can be seen from Fig.
\ref{f:rsf_dep_regularization} that the smallest value of the MRSE is achieved
for very small values of $\gamma$. This implies that the introduction of KS
bounds leads to the best results. The same can be seen from Table
\ref{t:rsf_synth_KS}, where the RSE measures under different conditions of the
experiment are shown for $10$ training examples. The measure $E_{3}$ is
obtained under condition $\gamma=0.005$. It can be seen from Table
\ref{t:cox_synth_KS} that, in contrast to the previous experiments (with the
black-box Cox model), the studied part of experiments shows outperforming
results with the KS bounds for most examples ($7$ from $10$).%

%TCIMACRO{\FRAME{ftbpFU}{4.1451in}{2.4915in}{0pt}{\Qcb{The MRSE as a function
%of hyper-parameter $\lambda$ and probability $\gamma$ for the RSF}%
%}{\Qlb{f:rsf_dep_regularization}}{rsf_dep_regularization.png}%
%width 4.1451in;  height 2.4915in;  depth 0pt;  original-width 9.9998in;
%original-height 6.0001in;  cropleft "0";  croptop "1";  cropright "1";
%filename 'rsf_dep_regularization.png';file-properties "XNPEU";}} }%
%BeginExpansion
%EndExpansion
%

%TCIMACRO{\TeXButton{B}{\begin{table}[tbp] \centering}}%
%BeginExpansion
\begin{table}[tbp] \centering
%EndExpansion
\caption{MRSE and RSE for three cases of using KS bounds with the large and small datasets for training  the RSF}%
\begin{tabular}
[c]{cccc}\hline
& \multicolumn{3}{c}{datasets}\\\hline
& large & \multicolumn{2}{c}{small}\\\hline
examples & $E_{1}$ & $E_{2}$ & $E_{3}$\\\hline
0 & $0.480$ & $0.031$ & $0.047$\\\hline
1 & $0.554$ & $0.139$ & $\mathbf{0.073}$\\\hline
2 & $0.627$ & $0.042$ & $0.078$\\\hline
3 & $0.146$ & $0.082$ & $\mathbf{0.045}$\\\hline
4 & $0.286$ & $0.142$ & $\mathbf{0.094}$\\\hline
5 & $0.231$ & $0.062$ & $\mathbf{0.042}$\\\hline
6 & $0.482$ & $0.051$ & $0.128$\\\hline
7 & $0.629$ & $0.160$ & $\mathbf{0.099}$\\\hline
8 & $0.142$ & $0.087$ & $\mathbf{0.074}$\\\hline
9 & $0.194$ & $0.150$ & $\mathbf{0.052}$\\\hline
$MRSE$ & $0.377$ & $0.095$ & $\mathbf{0.073}$\\\hline
\end{tabular}
\label{t:rsf_synth_KS}%
%TCIMACRO{\TeXButton{E}{\end{table}}}%
%BeginExpansion
\end{table}%
%EndExpansion

\begin{figure}
[ptb]
\begin{center}
\includegraphics[
%%=5.000400in,
%%=19.999599in,
height=1.4875in,
width=5.6844in
]%
{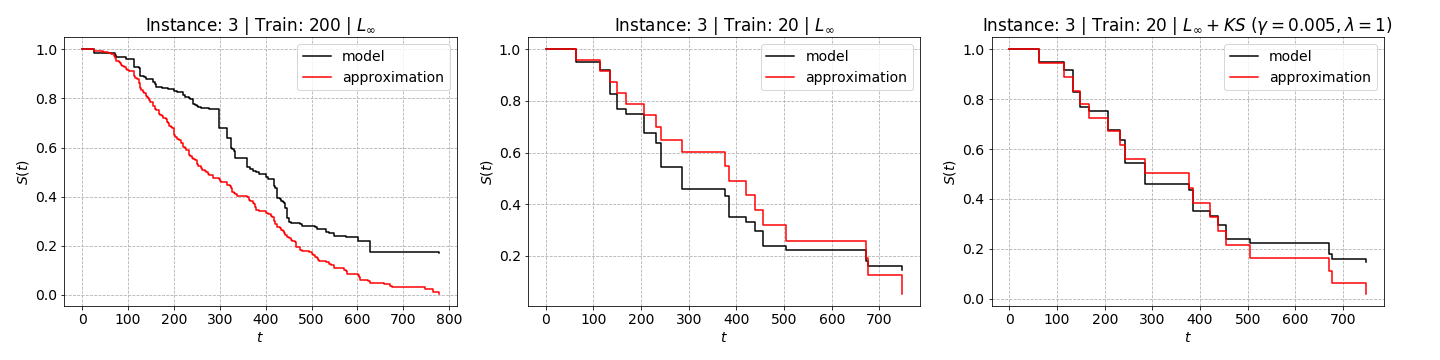}%
\caption{An illustration of the approximation results under three conditions
of experiments for the RSF by considering a \textquotedblleft
good\textquotedblright\ example}%
\label{fig:rsf_instance_3}%
\end{center}
\end{figure}

\begin{figure}
[ptb]
\begin{center}
\includegraphics[
%%=5.000400in,
%%=19.999599in,
height=1.4892in,
width=5.6931in
]%
{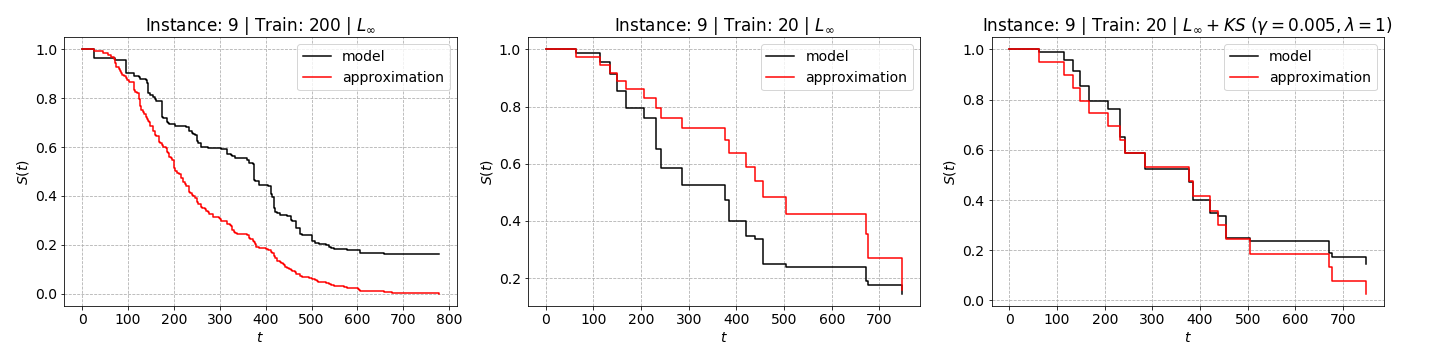}%
\caption{An illustration of the approximation results under three conditions
of experiments for the RSF by considering a \textquotedblleft
good\textquotedblright\ example}%
\label{fig:rsf_instance_9}%
\end{center}
\end{figure}

\begin{figure}
[ptb]
\begin{center}
\includegraphics[
%%=5.000400in,
%%=19.999599in,
height=1.4901in,
width=5.7104in
]%
{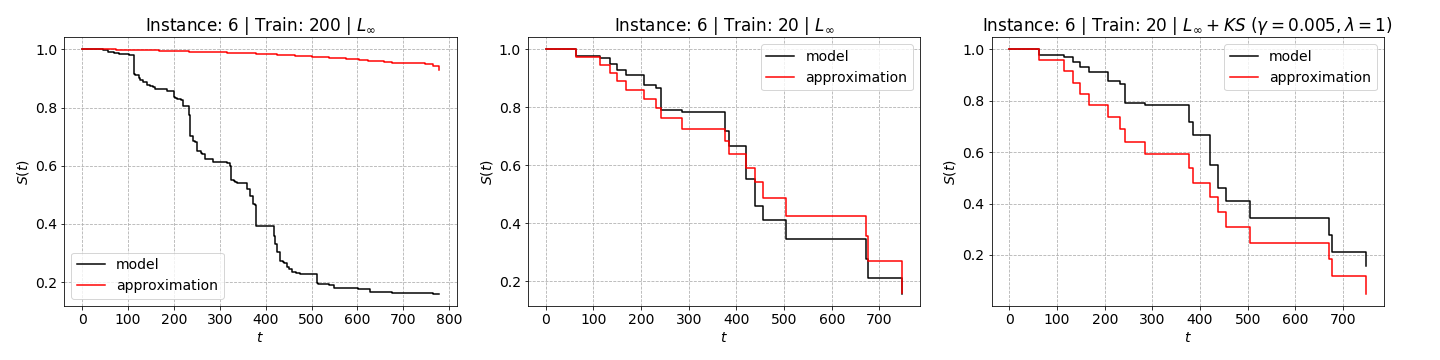}%
\caption{An illustration of the approximation results under three conditions
of experiments for the RSF by considering a \textquotedblleft
bad\textquotedblright\ example}%
\label{fig:rsf_instance_6}%
\end{center}
\end{figure}

The corresponding results are illustrated in Figs. \ref{fig:rsf_instance_3}%
-\ref{fig:rsf_instance_6}. In particular, Fig. \ref{fig:rsf_instance_3} shows
three considered cases of experiments. Values of coefficients $\mathbf{b}$ of
important features are not depicted for the RSF because the RSF does not
provide the important features like the Cox model. All figures illustrate
pairs of SFs obtained by means of the RSF and by means of the approximating
Cox model. Figs. \ref{fig:rsf_instance_3} and \ref{fig:rsf_instance_9} are
examples of a \textquotedblleft good\textquotedblright\ case when the
introduction of KS bounds leads to the better approximation and explanation.
In contrast to the examples in Figs. \ref{fig:rsf_instance_3} and
\ref{fig:rsf_instance_9}, Fig. \ref{fig:rsf_instance_6} illustrates a
\textquotedblleft bad\textquotedblright\ case when the introduction of KS
bounds leads to the worse approximation and explanation.%

%TCIMACRO{\FRAME{ftbpFU}{5.6844in}{1.4875in}{0pt}{\Qcb{An illustration of the
%approximation results under three conditions of experiments for the RSF by
%considering a \textquotedblleft good\textquotedblright\ example}%
%}{\Qlb{fig:rsf_instance_3}}{rsf_instance_3.png}%
%{\special{ language "Scientific Word";  type "GRAPHIC";  display "USEDEF";
%valid_file "F";  width 5.6844in;  height 1.4875in;  depth 0pt;
%original-width 19.9996in;  original-height 5.0004in;  cropleft "0";
%filename 'rsf_instance_3.png';file-properties "XNPEU";}} }%
%BeginExpansion
%EndExpansion
%

%TCIMACRO{\FRAME{ftbpFU}{5.6931in}{1.4892in}{0pt}{\Qcb{An illustration of the
%approximation results under three conditions of experiments for the RSF by
%considering a \textquotedblleft good\textquotedblright\ example}%
%}{\Qlb{fig:rsf_instance_9}}{rsf_instance_9.png}%
%{\special{ language "Scientific Word";  type "GRAPHIC";  display "USEDEF";
%valid_file "F";  width 5.6931in;  height 1.4892in;  depth 0pt;
%original-width 19.9996in;  original-height 5.0004in;  cropleft "0";
%filename 'rsf_instance_9.png';file-properties "XNPEU";}} }%
%BeginExpansion
%EndExpansion
%

%TCIMACRO{\FRAME{ftbpFU}{5.7104in}{1.4901in}{0pt}{\Qcb{An illustration of the
%approximation results under three conditions of experiments for the RSF by
%considering a \textquotedblleft bad\textquotedblright\ example}}%
%{\Qlb{fig:rsf_instance_6}}{rsf_instance_6.png}%
%{\special{ language "Scientific Word";  type "GRAPHIC";  display "USEDEF";
%valid_file "F";  width 5.7104in;  height 1.4901in;  depth 0pt;
%original-width 19.9996in;  original-height 5.0004in;  cropleft "0";
%filename 'rsf_instance_6.png';file-properties "XNPEU";}} }%
%BeginExpansion
%EndExpansion

\subsubsection{Contamination case}

\begin{figure}
[ptb]
\begin{center}
\includegraphics[
%%=1.499600in,
%%=1.499600in,
height=2.7337in,
width=2.7337in
]%
{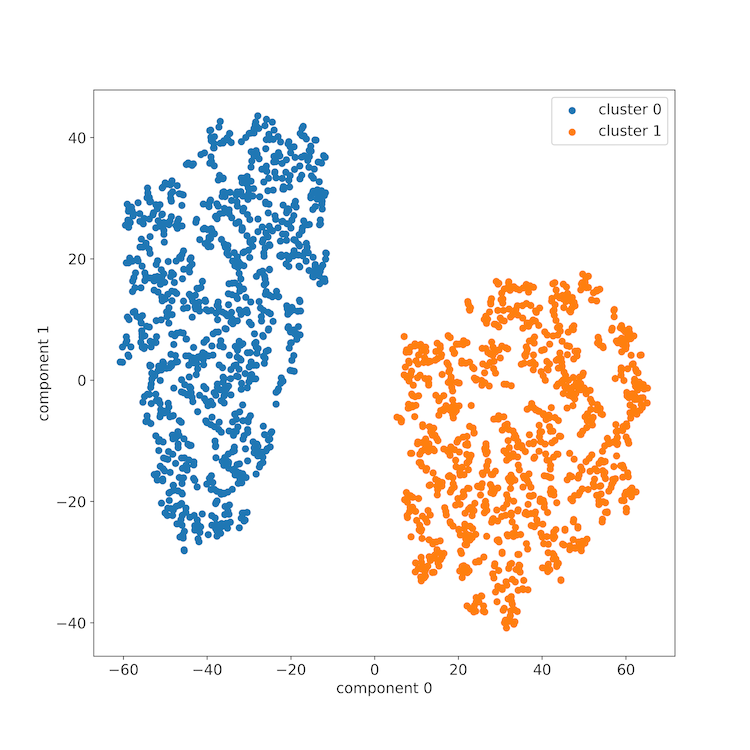}%
\caption{Two clusters of generated covariates depicted by using the t-SNE
method}%
\label{fig:two_cluster_tsne-KS}%
\end{center}
\end{figure}

The next part of numerical experiments aims to study how the proposed
explanation model deals with contaminated data, i.e., the objective is to get
the black-box model explanation which is robust to contaminated data. In order
to describe or generate so-called outliers in univariate statistical survival
data, we assume that some subset of examples from all training examples is
shifted, whereas other examples still come from some common target
distribution. Two clusters of covariates $\mathbf{x}\in\mathbb{R}^{d}$ are
randomly generated such that points of every cluster are generated from the
uniform distribution in the unit sphere. Centers of clusters are
$p_{0}=(2,2,2,2,2)$ and $p_{1}=(5,5,5,5,5)$, respectively. Parameters of the
generated clusters are chosen such that the clusters do not intersect each
other. The second cluster with the center $p_{1}$ is viewed as contaminated
data. The number of points in every cluster is $1000$. They are depicted in
Fig. \ref{fig:two_cluster_tsne-KS} using the well-known t-SNE algorithm.

\begin{figure}
[ptb]
\begin{center}
\includegraphics[
%%=2.000300in,
%%=6.000100in,
height=1.8853in,
width=4.4105in
]%
{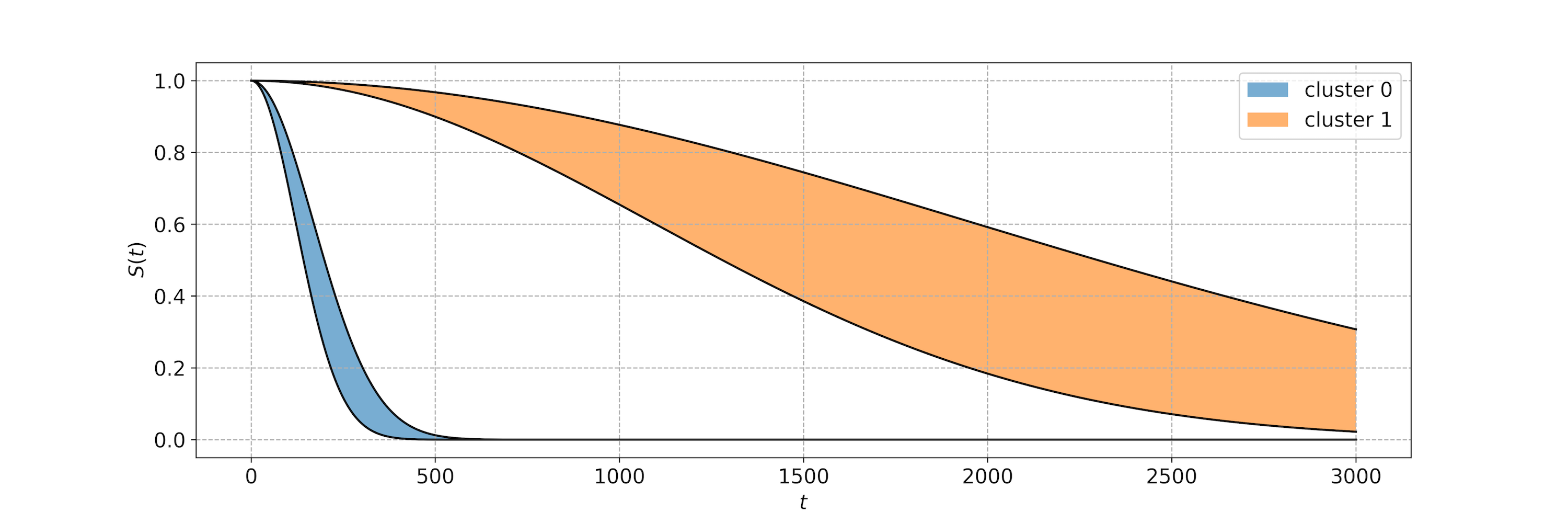}%
\caption{Sets of SFs corresponding to generated data for clusters 0 and 1}%
\label{fig:SFs_gen_KS}%
\end{center}
\end{figure}

Times to event are generated by using (\ref{KS_Inf_SurvLIME_84}) with
parameters $\lambda_{0}=10^{-5}$, $v=2$, $\mathbf{b}_{0,\text{true}}%
=(10^{-6},0.1,0.35,10^{-6},10^{-6})$ and $\mathbf{b}_{1,\text{true}}%
=(10^{-6},-0.6,10^{-6},10^{-6},0-0.15)$, where $\mathbf{b}_{0,\text{true}}$
and $\mathbf{b}_{1,\text{true}}$ are coefficient of the Cox model for
generating random points from clusters $0$ and $1$, respectively. The
parameters are taken in a way to get distinguished sets of SFs for both
clusters as it is shown in Fig. \ref{fig:SFs_gen_KS}, where two areas of SFs
are depicted corresponding to different clusters.%

%TCIMACRO{\FRAME{ftbpFU}{2.7337in}{2.7337in}{0pt}{\Qcb{Two clusters of
%generated covariates depicted by using the t-SNE method}}%
%{\Qlb{fig:two_cluster_tsne-KS}}{t_sne_ks.png}%
%width 2.7337in;  height 2.7337in;  depth 0pt;  original-width 1.4996in;
%original-height 1.4996in;  cropleft "0";  croptop "1";  cropright "1";
%cropbottom "0";  filename 't_SNE_KS.png';file-properties "XNPEU";}} }%
%BeginExpansion
%EndExpansion
%

%TCIMACRO{\FRAME{ftbpFU}{4.4105in}{1.8853in}{0pt}{\Qcb{Sets of SFs
%corresponding to generated data for clusters 0 and 1}}{\Qlb{fig:SFs_gen_KS}%
%}{surv_func_distribution.png}{\special{ language "Scientific Word";
%type "GRAPHIC";  display "USEDEF";  valid_file "F";  width 4.4105in;
%height 1.8853in;  depth 0pt;  original-width 6.0001in;
%original-height 2.0003in;  cropleft "0";  croptop "1";  cropright "1";
%filename 'surv_func_distribution.png';file-properties "XNPEU";}} }%
%BeginExpansion
%EndExpansion

Two cases are studied: the black-box model is trained on the large dataset
($500$ examples) and on the small dataset ($20$ examples). The model is tested
on $100$ examples.

The basic (uncontaminated) large training set is formed from randomly selected
$500$ examples of cluster $0$. The basic (uncontaminated) small training set
is formed from randomly selected $20$ examples of the same cluster. The
contaminated dataset is formed as follows. We find the largest time to event
$t_{\max}^{(0)}$ from uncontaminated examples. The first quarter of the
uncontaminated set consisting of $125$ and $5$ examples from the large and
small datasets, respectively, is selected. These examples are replaced with
examples from cluster 1 having times to event larger than $t_{\max}^{(0)}$. As
a result, we have a mix of two clusters. For numerical experiments, we
consider two methods: without KS bounds and with KS bounds. In sum, we get
four cases for studying:

Case 1. The explanation model without KS bounds is trained on the
uncontaminated dataset.

Case 2. The explanation model with KS bounds is trained on the uncontaminated dataset.

Case 3. The explanation model without KS bounds is trained on the contaminated dataset.

Case 4. The explanation model with KS bounds is trained on the contaminated dataset.

Measures $MRSE(H_{\text{model}},H_{\text{approx}})$ are computed for Cases 1-4
denoted as $S_{1}$, $S_{2}$, $S_{3}$, $S_{4}$, respectively, by testing $100$
examples. The results of numerical experiments corresponding to all the
considered cases for the black-box Cox model and the RSF are shown in Table
\ref{t:contam_synth_KS} where the best performance on each dataset is shown in
bold. It can be seen from Table \ref{t:contam_synth_KS} that three cases from
four ones of the contaminated dataset show outperforming results by using KS
bounds. Moreover, when the studied black-box model is the RSF, then the use of
KS bounds in all cases leads to better results.%

%TCIMACRO{\TeXButton{B}{\begin{table}[tbp] \centering}}%
%BeginExpansion
\begin{table}[tbp] \centering
%EndExpansion
\caption{MRSE for three cases of using KS bounds with the large and small datasets for training  the RSF}%
\begin{tabular}
[c]{cccccc}\hline
dataset &  & \multicolumn{2}{c}{uncontaminated} &
\multicolumn{2}{c}{contaminated}\\\hline
&  & without KS & with KS & without KS & with KS\\\hline
& model & $S_{1}$ & $S_{2}$ & $S_{3}$ & $S_{4}$\\\hline
large & Cox & $\mathbf{0.0085}$ & $0.0098$ & $0.3125$ & $\mathbf{0.1946}%
$\\\hline
large & RSF & $0.4130$ & $\mathbf{0.0760}$ & $0.4262$ & $\mathbf{0.0796}%
$\\\hline
small & Cox & $\mathbf{0.0239}$ & $0.0300$ & $\mathbf{0.3689}$ &
$0.3899$\\\hline
small & RSF & $0.0764$ & $\mathbf{0.0523}$ & $0.1258$ & $\mathbf{0.0796}%
$\\\hline
\end{tabular}
\label{t:contam_synth_KS}%
%TCIMACRO{\TeXButton{E}{\end{table}}}%
%BeginExpansion
\end{table}%
%EndExpansion

\begin{figure}
[ptb]
\begin{center}
\includegraphics[
%%=1.919900in,
%%=3.840600in,
height=2.3947in,
width=4.7573in
]%
{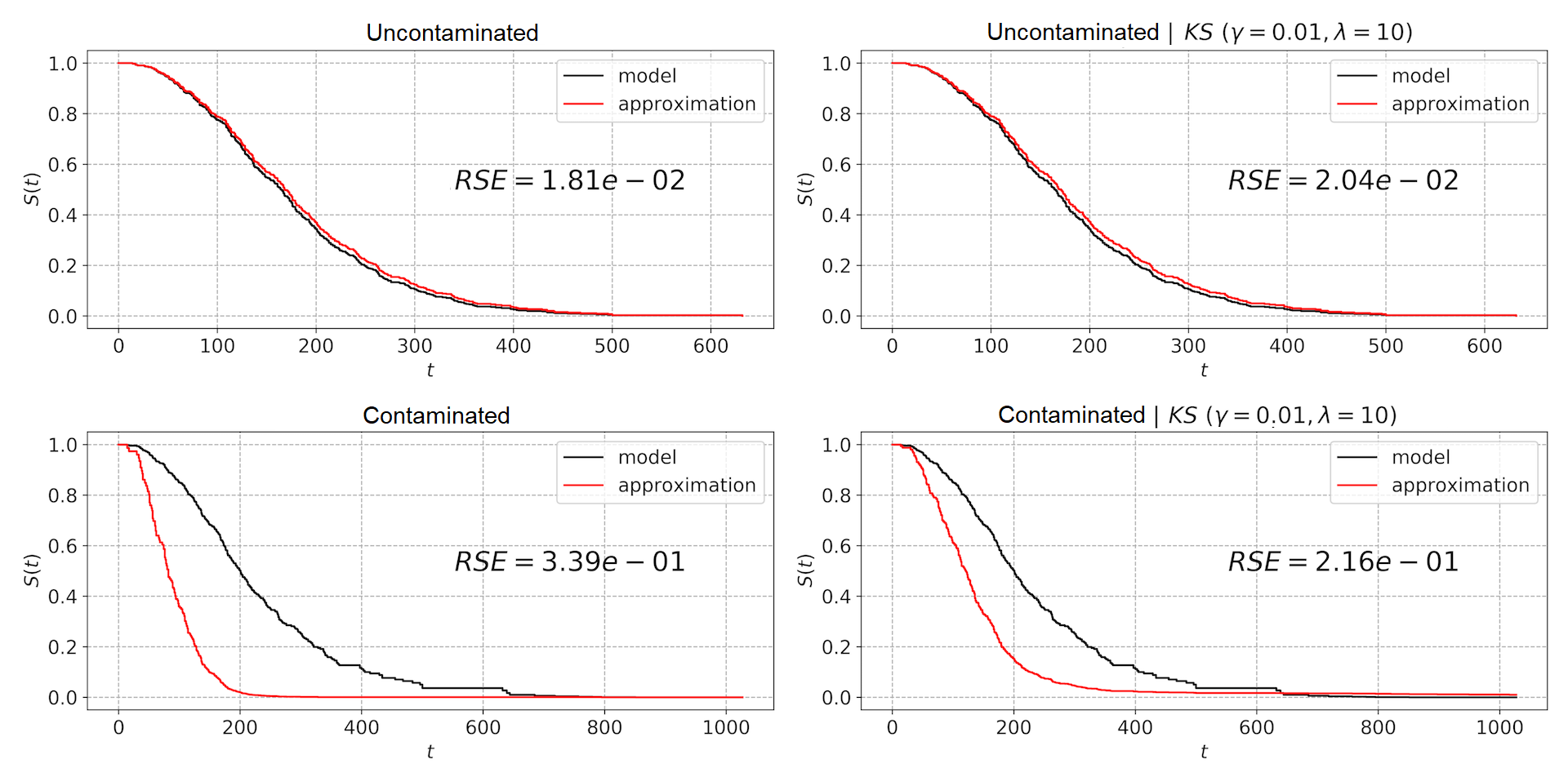}%
\caption{SFs of the black-box Cox model and the approximating Cox model for
two cases of large datasets without and with KS bounds}%
\label{fig:big_cox_figure}%
\end{center}
\end{figure}

\begin{figure}
[ptb]
\begin{center}
\includegraphics[
%%=1.919900in,
%%=3.840600in,
height=2.4025in,
width=4.7746in
]%
{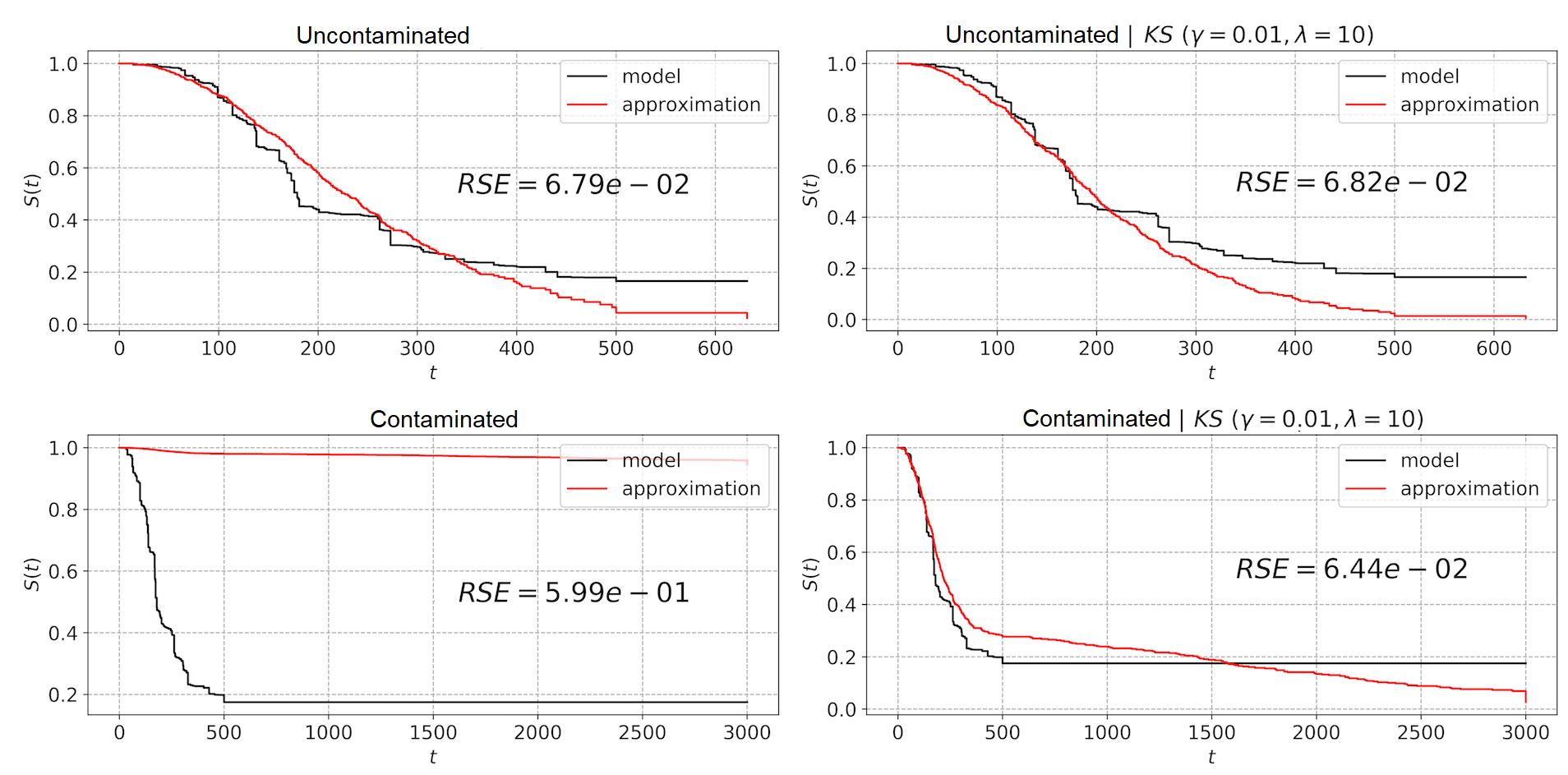}%
\caption{SFs of the RSF and the approximating Cox model for two cases of large
datasets without and with KS bounds}%
\label{fig:big_rsf_figure}%
\end{center}
\end{figure}

\begin{figure}
[ptb]
\begin{center}
\includegraphics[
%%=1.919900in,
%%=3.840600in,
height=2.4223in,
width=4.8092in
]%
{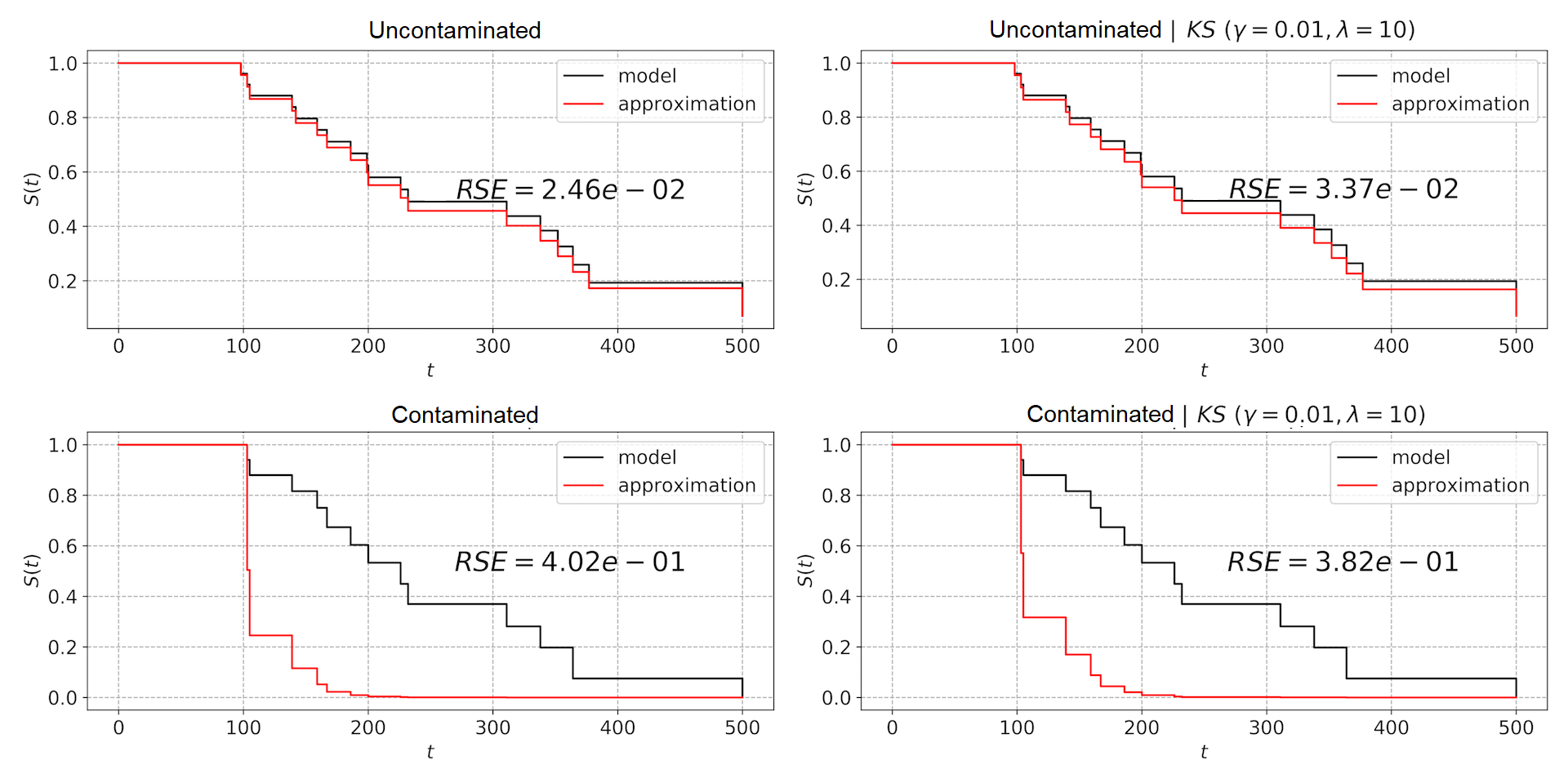}%
\caption{SFs of the black-box Cox model and the approximating Cox model for
two cases of small datasets without and with KS bounds}%
\label{fig:small_cox_figure}%
\end{center}
\end{figure}

\begin{figure}
[ptb]
\begin{center}
\includegraphics[
%%=1.919900in,
%%=3.840600in,
height=2.4267in,
width=4.8179in
]%
{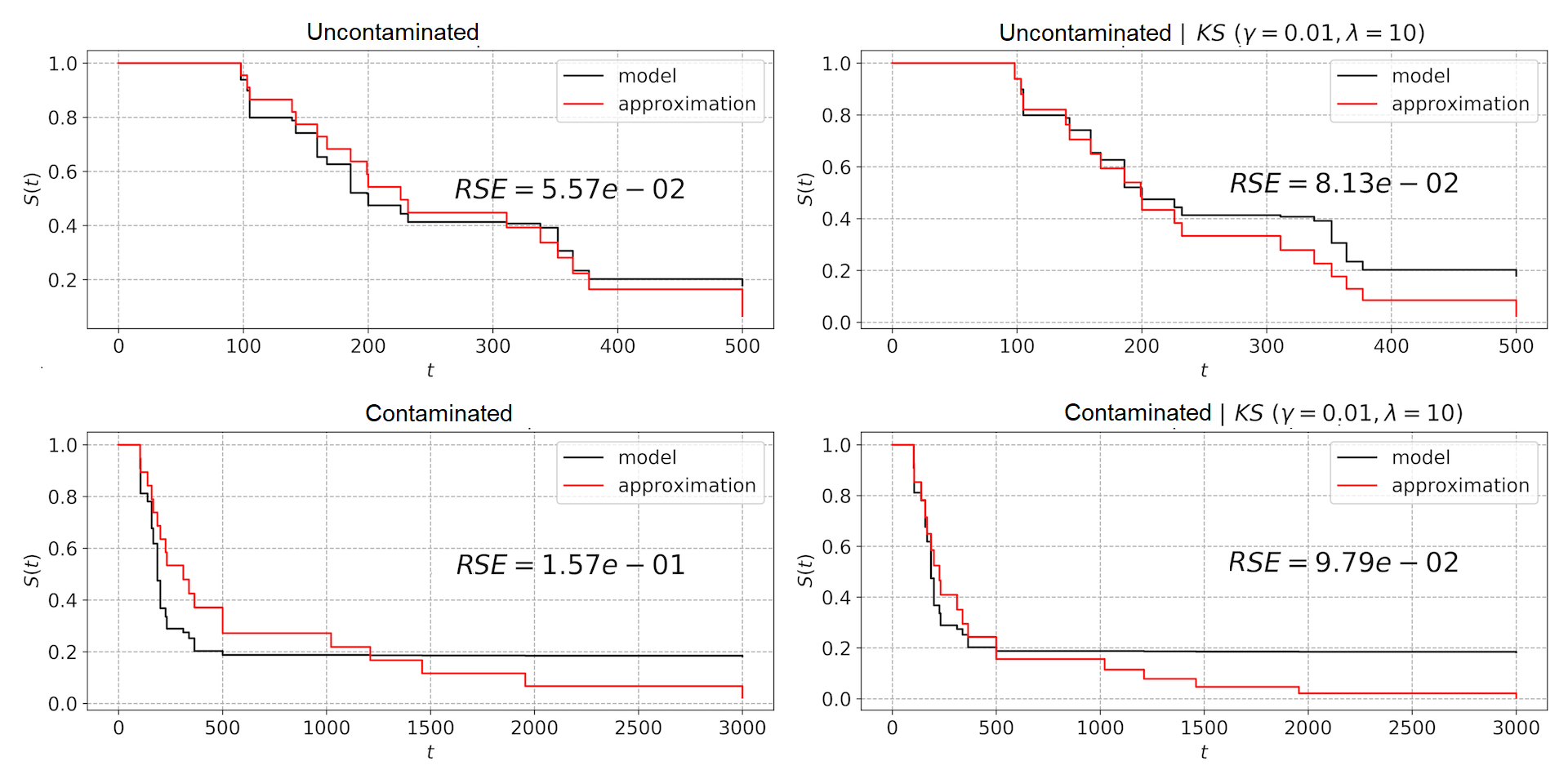}%
\caption{SFs of the RSF and the approximating Cox model for two cases of small
datasets without and with KS bounds}%
\label{fig:small_rsf_figure}%
\end{center}
\end{figure}

In order to illustrate the difference between the results, we show the SFs in
Figs. \ref{fig:big_cox_figure}-\ref{fig:small_rsf_figure}. Every figure
consists of four pictures illustrating a pair of the black-box model SF and
the approximating SF. Pictures in the first (the second) row show SFs of
models trained on the uncontaminated (contaminated) dataset. Pictures in the
first (the second) column illustrate SFs obtained without (with) using KS
bounds. The same pictures can be found in Figs. \ref{fig:big_rsf_figure}%
-\ref{fig:small_rsf_figure}. Results in Figs. \ref{fig:big_cox_figure}%
-\ref{fig:small_rsf_figure} confirm conclusions which have been made by
analyzing Table \ref{t:contam_synth_KS}. In the case of uncontaminated
datasets, the use of KS bounds gives comparable approximations except for the
case of the RSF trained on the small dataset (see Fig.
\ref{fig:small_rsf_figure}). At the same time, one can see quite different
results when the black-box models are trained on the contaminated datasets
where the use of KS bounds provides outperforming or comparable
approximations. Moreover, the quality of approximation increases when the
models are trained on the small dataset.%

%TCIMACRO{\FRAME{ftbpFU}{4.7573in}{2.3947in}{0pt}{\Qcb{SFs of the black-box Cox
%model and the approximating Cox model for two cases of large datasets without
%and with KS bounds}}{\Qlb{fig:big_cox_figure}}{big_cox_figure.png}%
%width 4.7573in;  height 2.3947in;  depth 0pt;  original-width 3.8406in;
%original-height 1.9199in;  cropleft "0";  croptop "1";  cropright "1";
%cropbottom "0";  filename 'big_cox_figure.png';file-properties "XNPEU";}} }%
%BeginExpansion
%EndExpansion
%

%TCIMACRO{\FRAME{ftbpFU}{4.7746in}{2.4025in}{0pt}{\Qcb{SFs of the RSF and the
%approximating Cox model for two cases of large datasets without and with KS
%bounds}}{\Qlb{fig:big_rsf_figure}}{big_rsf_figure.png}%
%width 4.7746in;  height 2.4025in;  depth 0pt;  original-width 3.8406in;
%original-height 1.9199in;  cropleft "0";  croptop "1";  cropright "1";
%cropbottom "0";  filename 'big_rsf_figure.png';file-properties "XNPEU";}} }%
%BeginExpansion
%EndExpansion
%

%TCIMACRO{\FRAME{ftbpFU}{4.8092in}{2.4223in}{0pt}{\Qcb{SFs of the black-box Cox
%model and the approximating Cox model for two cases of small datasets without
%and with KS bounds}}{\Qlb{fig:small_cox_figure}}{small_cox_figure.png}%
%width 4.8092in;  height 2.4223in;  depth 0pt;  original-width 3.8406in;
%original-height 1.9199in;  cropleft "0";  croptop "1";  cropright "1";
%cropbottom "0";  filename 'small_cox_figure.png';file-properties "XNPEU";}} }%
%BeginExpansion
%EndExpansion
%

%TCIMACRO{\FRAME{ftbpFU}{4.8179in}{2.4267in}{0pt}{\Qcb{SFs of the RSF and the
%approximating Cox model for two cases of small datasets without and with KS
%bounds}}{\Qlb{fig:small_rsf_figure}}{small_rsf_figure.png}%
%width 4.8179in;  height 2.4267in;  depth 0pt;  original-width 3.8406in;
%original-height 1.9199in;  cropleft "0";  croptop "1";  cropright "1";
%cropbottom "0";  filename 'small_rsf_figure.png';file-properties "XNPEU";}} }%
%BeginExpansion
%EndExpansion

\subsection{Numerical experiments with real data}

We consider the following well-known real datasets to study SurvLIME-KS.

The Veterans' Administration Lung Cancer Study (Veteran) Dataset
\cite{Kalbfleisch-Prentice-1980} contains data on 137 males with advanced
inoperable lung cancer. The subjects were randomly assigned to either a
standard chemotherapy treatment or a test chemotherapy treatment. Several
additional variables were also measured on the subjects. The number of
features is 6, but it is extended till 9 due to categorical features.

The NCCTG Lung Cancer (LUNG) Dataset \cite{Loprinzi-etal-1994} records the
survival of patients with advanced lung cancer, together with assessments of
the patients performance status measured either by the physician and by the
patients themselves. The data set contains 228 patients, including 63 patients
that are right censored (patients that left the study before their death). The
number of features is 8, but it is extended till 11 due to categorical features.

The above datasets can be downloaded via R package \textquotedblleft
survival\textquotedblright.

Approximation accuracy measures ($E_{1}$, $E_{2}$, $E_{3}$) are obtained for
three cases: 1) approximating without KS bounds on the large dataset; 2)
approximating without KS bounds on the small dataset; 3) approximating with KS
bounds on the small dataset. The corresponding results for the black-box Cox
model and the RSF trained on the Veteran dataset are shown in Table
\ref{t:veteran_cox_rsf}. It can be seen from Table \ref{t:veteran_cox_rsf}
that the Cox model provides better results without KS bounds for the large
dataset as well as the small dataset. The RSF shows quite different results.
One can see that the introduction of KS bounds gives better result shown in
bold ($8$ from $10$). We again show in bold only results corresponding to the
use of KS bounds.%

%TCIMACRO{\TeXButton{B}{\begin{table}[tbp] \centering}}%
%BeginExpansion
\begin{table}[tbp] \centering
%EndExpansion
\caption{RSE and MRSE for three cases of using KS bounds with the large and small data obtained from the Veteran dataset}%
\begin{tabular}
[c]{ccccccc}\hline
& \multicolumn{3}{c}{Cox model} & \multicolumn{3}{c}{RSF}\\\hline
dataset & large & \multicolumn{2}{c}{small} & large &
\multicolumn{2}{c}{small}\\\hline
& $E_{1}$ & $E_{2}$ & $E_{3}$ & $E_{1}$ & $E_{2}$ & $E_{3}$\\\hline
0 & $0.044$ & $0.089$ & $0.159$ & $0.154$ & $0.044$ & $0.046$\\\hline
1 & $0.083$ & $0.045$ & $0.094$ & $0.150$ & $0.037$ & $0.040$\\\hline
2 & $0.052$ & $0.036$ & $0.053$ & $0.126$ & $0.072$ & $\mathbf{0.058}$\\\hline
3 & $0.046$ & $0.077$ & $0.097$ & $0.187$ & $0.112$ & $\mathbf{0.073}$\\\hline
4 & $0.061$ & $0.045$ & $0.067$ & $0.643$ & $0.085$ & $\mathbf{0.059}$\\\hline
5 & $0.031$ & $0.033$ & $0.044$ & $0.167$ & $0.098$ & $\mathbf{0.093}$\\\hline
6 & $0.051$ & $0.047$ & $0.077$ & $0.649$ & $0.073$ & $\mathbf{0.057}$\\\hline
7 & $0.045$ & $0.034$ & $0.039$ & $0.605$ & $0.063$ & $\mathbf{0.048}$\\\hline
8 & $0.047$ & $0.037$ & $0.038$ & $0.607$ & $0.041$ & $\mathbf{0.040}$\\\hline
9 & $0.032$ & $0.073$ & $0.093$ & $0.218$ & $0.115$ & $\mathbf{0.059}$\\\hline
$MRSE$ & $0.049$ & $0.052$ & $0.076$ & $0.351$ & $0.074$ & $\mathbf{0.057}%
$\\\hline
\end{tabular}
\label{t:veteran_cox_rsf}%
%TCIMACRO{\TeXButton{E}{\end{table}}}%
%BeginExpansion
\end{table}%
%EndExpansion

Similar results for the LUNG dataset are given in Table \ref{t:lung_cox_rsf}.
The use of KS bounds for this dataset shows even better results than for the
Veteran dataset. One can see from Table \ref{t:lung_cox_rsf} that
outperforming results are available for the small dataset with using the Cox model.%

%TCIMACRO{\TeXButton{B}{\begin{table}[tbp] \centering}}%
%BeginExpansion
\begin{table}[tbp] \centering
%EndExpansion
\caption{RSE and MRSE for three cases of using KS bounds with the large and small data obtained from the Veteran dataset}%
\begin{tabular}
[c]{ccccccc}\hline
& \multicolumn{3}{c}{Cox model} & \multicolumn{3}{c}{RSF}\\\hline
dataset & large & \multicolumn{2}{c}{small} & large &
\multicolumn{2}{c}{small}\\\hline
& $E_{1}$ & $E_{2}$ & $E_{3}$ & $E_{1}$ & $E_{2}$ & $E_{3}$\\\hline
0 & $0.061$ & $0.036$ & $0.157$ & $0.093$ & $0.088$ & $\mathbf{0.055}$\\\hline
1 & $0.079$ & $0.025$ & $0.089$ & $0.098$ & $0.054$ & $0.077$\\\hline
2 & $0.027$ & $0.134$ & $0.253$ & $0.588$ & $0.076$ & $\mathbf{0.064}$\\\hline
3 & $0.039$ & $0.061$ & $\mathbf{0.036}$ & $0.419$ & $0.097$ & $\mathbf{0.069}%
$\\\hline
4 & $0.043$ & $0.184$ & $\mathbf{0.151}$ & $0.327$ & $0.085$ & $0.097$\\\hline
5 & $0.095$ & $0.031$ & $0.127$ & $0.083$ & $0.065$ & $\mathbf{0.062}$\\\hline
6 & $0.019$ & $0.158$ & $0.199$ & $0.118$ & $0.110$ & $\mathbf{0.077}$\\\hline
7 & $0.017$ & $0.174$ & $\mathbf{0.139}$ & $0.502$ & $0.104$ & $\mathbf{0.067}%
$\\\hline
8 & $0.031$ & $0.122$ & $0.232$ & $0.574$ & $0.086$ & $\mathbf{0.070}$\\\hline
9 & $0.047$ & $0.177$ & $\mathbf{0.175}$ & $0.316$ & $0.091$ & $\mathbf{0.083}%
$\\\hline
$MRSE$ & $0.046$ & $0.110$ & $0.156$ & $0.312$ & $0.086$ & $\mathbf{0.072}%
$\\\hline
\end{tabular}
\label{t:lung_cox_rsf}%
%TCIMACRO{\TeXButton{E}{\end{table}}}%
%BeginExpansion
\end{table}%
%EndExpansion

\begin{figure}
[ptb]
\begin{center}
\includegraphics[
%%=9.999800in,
%%=19.999599in,
height=3.032in,
width=6.0459in
]%
{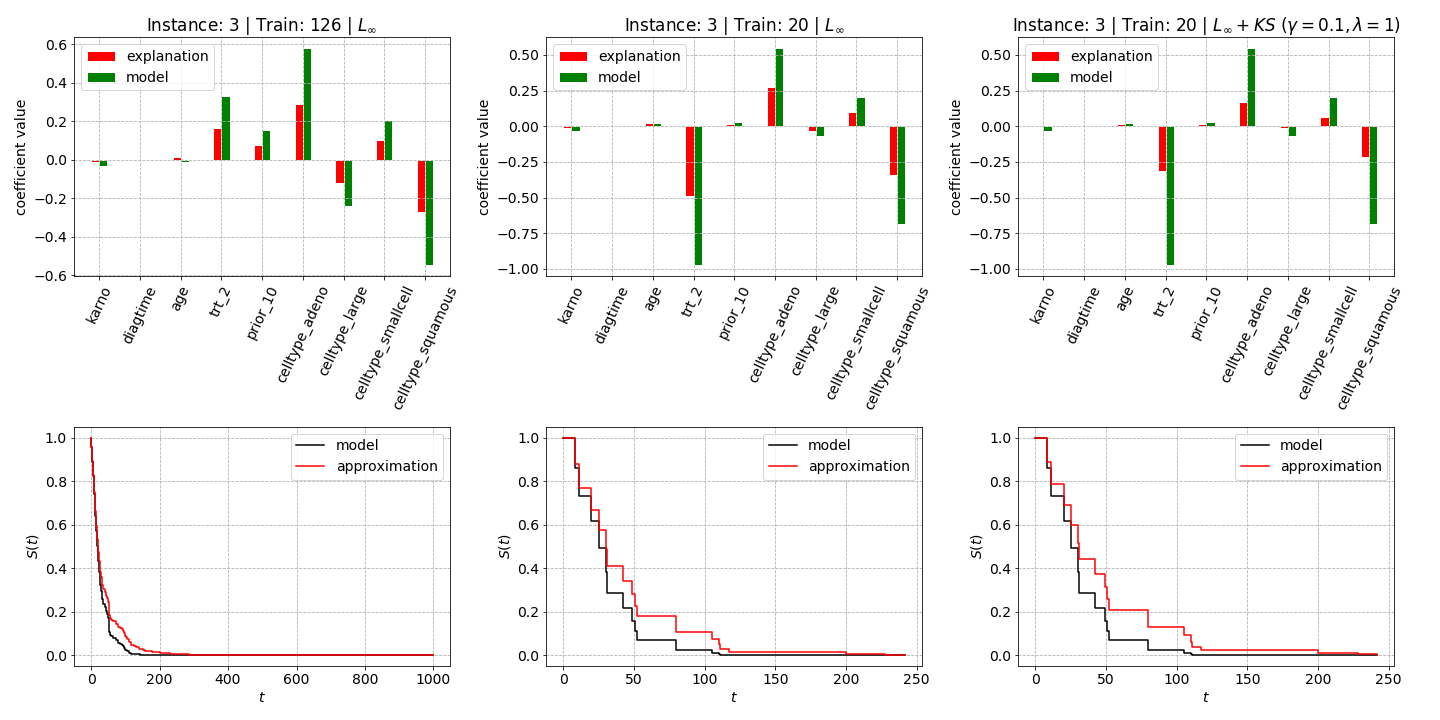}%
\caption{An illustration of the approximation results under three conditions
of experiments for the black-box Cox model trained on the Veteran dataset}%
\label{fig:veteran_cox_instance_3}%
\end{center}
\end{figure}

\begin{figure}
[ptb]
\begin{center}
\includegraphics[
%%=5.208700in,
%%=20.832800in,
height=1.5506in,
width=6.1402in
]%
{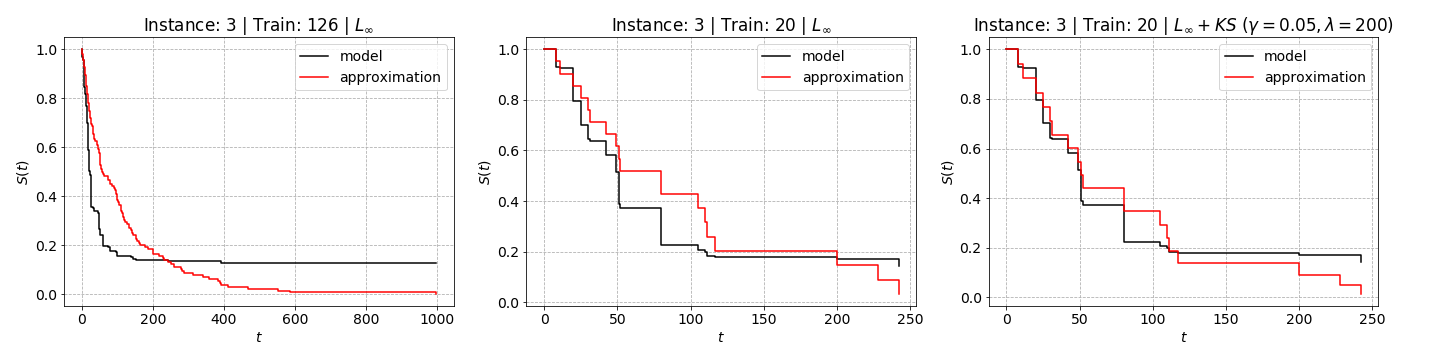}%
\caption{An illustration of the approximation results under three conditions
of experiments for the RSF trained on the Veteran dataset}%
\label{fig:veteran_rsf_instance_3}%
\end{center}
\end{figure}

Fig. \ref{fig:veteran_cox_instance_3} illustrates numerical results for the
black-box Cox model trained on the Veteran dataset again under three
conditions (approximating without KS bounds on the large dataset;
approximating without KS bounds on the small dataset; approximating with KS
bounds on the small dataset). The first row of pictures in Fig.
\ref{fig:veteran_cox_instance_3} illustrates the coefficients $\mathbf{b}$ of
important features for the three conditions (see similar pictures in Figs.
\ref{fig:cox_instance_2}-\ref{fig:cox_instance_3} for synthetic data). Two
vectors of features are depicted in the form of diagrams: $\mathbf{b}%
_{\text{model}}$ and $\mathbf{b}_{\text{expl}}$. The second row of pictures in
Fig. \ref{fig:veteran_cox_instance_3} illustrates how the SFs obtained by
means of the black-box Cox model (\textquotedblleft model\textquotedblright)
and by means of the approximating Cox model (\textquotedblleft
approximation\textquotedblright) are close to each other. The same results for
the RSF\ trained on the Veteran dataset are shown in Fig.
\ref{fig:veteran_rsf_instance_3}.

\begin{figure}
[ptb]
\begin{center}
\includegraphics[
%%=10.416400in,
%%=20.832800in,
height=2.9863in,
width=5.9518in
]%
{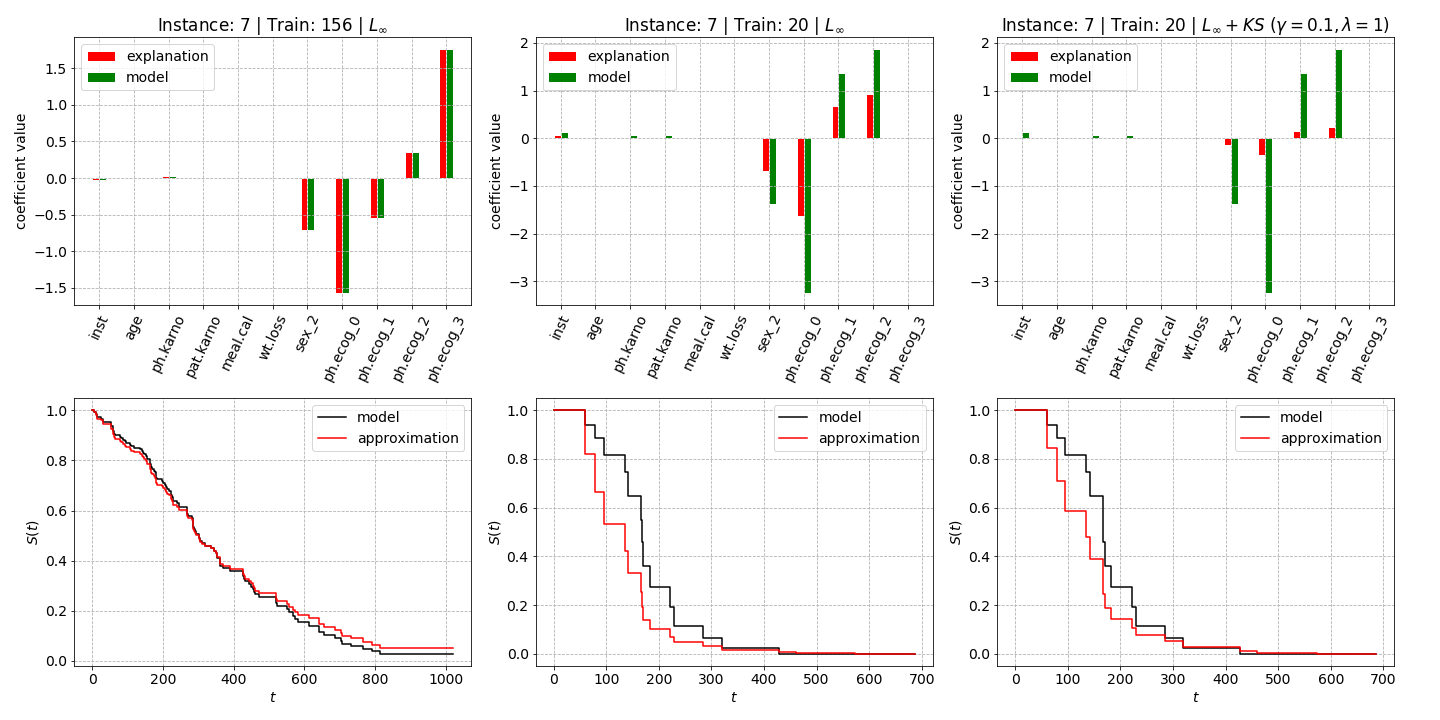}%
\caption{An illustration of the approximation results under three conditions
of experiments for the black-box Cox model trained on the LUNG dataset}%
\label{fig:lung_cox_instance_7}%
\end{center}
\end{figure}

\begin{figure}
[ptb]
\begin{center}
\includegraphics[
%%=5.208700in,
%%=20.832800in,
height=1.5215in,
width=6.0284in
]%
{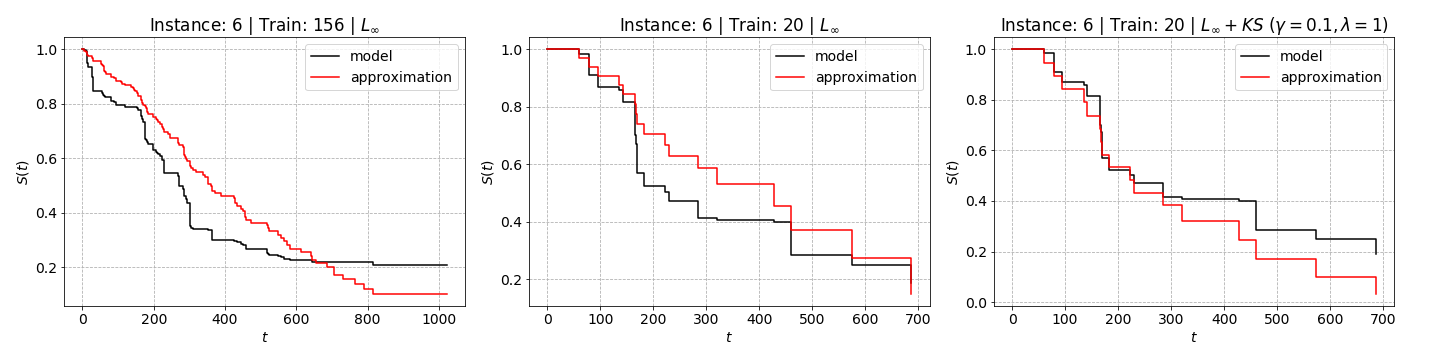}%
\caption{An illustration of the approximation results under three conditions
of experiments for the RSF trained on the LUNG dataset}%
\label{fig:lung_rsf_instance_6}%
\end{center}
\end{figure}

Similar numerical results for the LUNG dataset are shown in Figs.
\ref{fig:lung_cox_instance_7} and \ref{fig:lung_rsf_instance_6}.%

%TCIMACRO{\FRAME{ftbpFU}{6.0459in}{3.032in}{0pt}{\Qcb{An illustration of the
%approximation results under three conditions of experiments for the black-box
%Cox model trained on the Veteran dataset}}{\Qlb{fig:veteran_cox_instance_3}%
%}{veteran_cox_instance_3.png}{\special{ language "Scientific Word";
%type "GRAPHIC";  maintain-aspect-ratio TRUE;  display "USEDEF";
%valid_file "F";  width 6.0459in;  height 3.032in;  depth 0pt;
%original-width 19.9996in;  original-height 9.9998in;  cropleft "0";
%filename 'veteran_cox_instance_3.png';file-properties "XNPEU";}} }%
%BeginExpansion
%EndExpansion
%

%TCIMACRO{\FRAME{ftbpFU}{6.1402in}{1.5506in}{0pt}{\Qcb{An illustration of the
%approximation results under three conditions of experiments for the RSF
%trained on the Veteran dataset}}{\Qlb{fig:veteran_rsf_instance_3}%
%}{veteran_rsf_instance_3.png}{\special{ language "Scientific Word";
%type "GRAPHIC";  maintain-aspect-ratio TRUE;  display "USEDEF";
%valid_file "F";  width 6.1402in;  height 1.5506in;  depth 0pt;
%original-width 20.8328in;  original-height 5.2087in;  cropleft "0";
%filename 'veteran_rsf_instance_3.png';file-properties "XNPEU";}} }%
%BeginExpansion
%EndExpansion
%

%TCIMACRO{\FRAME{ftbpFU}{5.9518in}{2.9863in}{0pt}{\Qcb{An illustration of the
%approximation results under three conditions of experiments for the black-box
%Cox model trained on the LUNG dataset}}{\Qlb{fig:lung_cox_instance_7}%
%}{lung_cox_instance_7.png}{\special{ language "Scientific Word";
%type "GRAPHIC";  maintain-aspect-ratio TRUE;  display "USEDEF";
%valid_file "F";  width 5.9518in;  height 2.9863in;  depth 0pt;
%original-width 20.8328in;  original-height 10.4164in;  cropleft "0";
%filename 'lung_cox_instance_7.png';file-properties "XNPEU";}} }%
%BeginExpansion
%EndExpansion
%

%TCIMACRO{\FRAME{ftbpFU}{6.0284in}{1.5215in}{0pt}{\Qcb{An illustration of the
%approximation results under three conditions of experiments for the RSF
%trained on the LUNG dataset}}{\Qlb{fig:lung_rsf_instance_6}}%
%{lung_rsf_instance_6.png}{\special{ language "Scientific Word";
%type "GRAPHIC";  maintain-aspect-ratio TRUE;  display "USEDEF";
%valid_file "F";  width 6.0284in;  height 1.5215in;  depth 0pt;
%original-width 20.8328in;  original-height 5.2087in;  cropleft "0";
%filename 'lung_rsf_instance_6.png';file-properties "XNPEU";}} }%
%BeginExpansion
%EndExpansion

\section{Conclusion}

A new robust explanation algorithm called SurvLIME-KS which can be regarded as
a modification of the method SurvLIME has been presented in the paper. Its aim
is to get robust explanation under condition of small training data and
possible outliers. The basic idea behind the method is to approximate a set of
CHFs, which are predicted by the black-box survival model and are restricted
by KS bounds, by the CHF of the Cox proportional hazards model. The
approximating Cox model allows us to get important features explaining the
survival model. In contrast to SurvLIME, the proposed algorithm considers sets
of CHFs produced by the KS bounds.

Various numerical experiments with synthetic and real data have illustrated
the advantage of SurvLIME-KS in comparison with SurvLIME for certain
conditions on the training data.

It should be noted that we have used KS bounds which are rather conservative
bounds. However, there are other interesting bounds and imprecise statistical
models which also could be studied and incorporated into SurvLIME to get
robust algorithms, for example, the imprecise Dirichlet model \cite{Walley96a}%
, the imprecise pari-mutuel model, the linear-vacuous mixture or $\varepsilon
$-contaminated model, the constant odds-ratio model \cite{Walley91}. These
models may significantly improve the algorithm under some conditions. Their
use is another direction for further research.

We have studied only the case when KS bounds are used for predicted CHFs which
are produced by the black-box survival model. However, it is very interesting
to consider the same bounds or other bounds for CHFs produced by the
approximating Cox model. This case is more complicated, but it may enhance
robust properties of the algorithm. This is also a direction for further research.

\section*{Acknowledgement}

The reported study was funded by RFBR, project number 20-01-00154.

\bibliographystyle{plain}
\bibliography{Boosting,Classif_bib,Deep_Forest,Explain,Explain_med,IntervalClass,Medical,MYBIB,MYUSE,Robots,Survival_analysis}

\end{document}